\documentclass[letterpaper, 10 pt, conference]{ieeeconf}  
\pdfoutput=1

\IEEEoverridecommandlockouts                              

\usepackage{url}            
\usepackage{booktabs}       
\usepackage{amsfonts}       
\usepackage{nicefrac}       
\usepackage{microtype}      

\usepackage{epsfig}
\usepackage{graphicx}
\usepackage{amsmath}
\usepackage{amssymb}
\usepackage{subfigure}
\usepackage[table]{xcolor}
\usepackage{paralist}
\usepackage{booktabs}
\usepackage[]{algorithm2e}

\def\s{{\bf s}}

\def\v{{\bf v}}
\def\q{{\bf q}}
\def\w{{\bf w}}
\def\0{{\bf 0}}
\def\BE{{\mathbb E}}
\def\BR{{\mathbb R}}

\graphicspath{{fig/}}

\title{Learning Unmanned Aerial Vehicle Control for Autonomous Target Following}

\author{Siyi Li$^{1}$, Tianbo Liu$^{2}$, Chi Zhang$^{1}$, Dit-Yan Yeung$^{1}$, Shaojie Shen$^{2}$
\thanks{$^{1}$Department of Computer Science and Engineering, Hong Kong University of Science and Technology, Hong Kong, China.}
\thanks{$^{2}$Department of Electronic and Computer Engineering, Hong Kong University of Science and Technology, Hong Kong, China.}
}

\begin{document}

\maketitle
\thispagestyle{empty}
\pagestyle{empty}

\begin{abstract}
	While deep reinforcement learning (RL) methods have achieved unprecedented successes in a range of challenging problems, their applicability has been mainly limited to simulation or game domains due to the high sample complexity of the trial-and-error learning process. However, real-world robotic applications often need a data-efficient learning process with safety-critical constraints. In this paper, we consider the challenging problem of learning unmanned aerial vehicle (UAV) control for tracking a moving target. To acquire a strategy that combines perception and control, we represent the policy by a convolutional neural network. We develop a hierarchical approach that combines a model-free policy gradient method with a conventional feedback proportional-integral-derivative (PID) controller to enable stable learning without catastrophic failure. The neural network is trained by a combination of supervised learning from raw images and reinforcement learning from games of self-play. We show that the proposed approach can learn a target following policy in a simulator efficiently and the learned behavior can be successfully transferred to the DJI quadrotor platform for real-world UAV control. 
\end{abstract}

\section{Introduction}
The recent development of perception~\cite{drone-tracking,shen2013vision} and control~\cite{lee2011geometric}  technologies for unmanned aerial vehicles (UAVs) enables autonomy in various complex environments, opening up a promising market with applications in aerial photography, monitoring, and inspection. Many of these applications require an aerial robot to autonomously follow a moving target.

In this paper, we consider the problem of controlling a quadrotor with very limited payload to autonomously follow a moving target. Most early research on UAV autonomy focused on a two-step pipeline.
The first step is on perception where various vision algorithms are used to estimate the underlying state or map~\cite{rgbd-quadrotor,barry2015pushbroom,schmid2013stereo,bills2011autonomous,sun2017gesture}. The second step is to design and tune conventional feedback controllers, such as proportional-integral-derivative (PID) controllers, matching some manually defined rules.  
The whole system consists of several hard-coded components without any learning capability and relies heavily on tedious tuning by human experts to achieve good performance.

These limitations may be addressed by viewing the problem from the machine learning perspective. In essence, quadrotor control is a sequential prediction problem with the sensory information as input and the motor control commands as output. Both standard supervised learning~\cite{giusti2016machine} and imitation learning~\cite{uav-imitation} have been used to learn various controllers for UAVs. However, both methods require explicitly labeled datasets, which have to be provided by costly human experts. With human experts becoming the bottleneck, such approaches are often restricted to small datasets and thus cannot exploit high-capacity learning algorithms to train the policies.

Recent advances in reinforcement learning (RL) offer new promises for solving the problem. Instead of demanding explicitly labeled samples as in supervised learning, RL only requires a scalar reward function to guide the learning agent through a trial-and-error process interacting with the environment. Most RL approaches belong to either one of two categories:  model-based  methods and model-free methods. 
Model-based methods learn an explicit dynamical model of the environment and then optimize the policy under this model. They have been successfully applied to robotics for various applications such as object manipulation~\cite{pilco,deep-visuo}, ground vehicles~\cite{mueller2012iterative}, helicopters~\cite{helicopter,abbeel2007application}, and quadrotors~\cite{mohajerin2014modular,punjani2015deep,mpcgps}. Model-based methods tend to have a data-efficient learning process but suffer from significant bias since complex unknown dynamics cannot always be modeled accurately. 
In contrast, model-free methods have the ability of handling arbitrary dynamical systems with minimal bias. Several recent studies in model-free methods, especially deep RL~\cite{dqn,ddpg,a3c,trpo,gu2017deep}, have shown that end-to-end model-free methods are capable of learning high-quality control policies using generic neural networks with minimal feature engineering. However, learning deep neural networks for quadrotor controllers that map image pixels and states directly to low-level motor commands also poses a number of challenges. Model-free methods generally require a data-hungry training paradigm which is costly for real-world physical systems. Specific to the quadrotors, the relative frailty of the underlying system makes a partially trained motor-level policy crash in most of  the training process. A key question is how to enjoy the richness and flexibility of a self-improving policy by model-free learning while preserving the stability of conventional controllers.

In this paper, we propose to combine the stability of conventional feedback PID controllers with the self-improving performance of model-free RL techniques so that the hybrid method can be practically applied to learning UAV control. Model-free methods are used to learn the high-level reference inputs to the PID control loop while the PID controller maps the reference inputs to low-level motor actions. Consequently, both data efficiency and training stability will be greatly improved. Moreover, the proposed hierarchical control system makes it easy to transfer the policy trained in simulators to real-world platforms, since both sides share similar high-level system dynamics. This transfer ability is crucial to many real-world control tasks since learning in simulated environments only incurs low cost.
We represent both the perception module and the control module using a single convolutional neural network (CNN). The perception layers can be pre-trained by supervised learning which provides efficient learning updates. The whole network is then trained by the model-free RL approach. In so doing, it both adapts the perception module to optimize the task performance and adjusts the control module towards the goal of following the moving target. We demonstrate that the proposed approach can efficiently learn to control the quadrotor to follow a moving target under various simulated circumstances and the learned policy can be successfully transferred to a quadrotor platform for real-world control.


\section{Preliminaries}

\subsection{Formulation of Quadrotor Control Problem}
The quadrotor target following task can be naturally formulated as a sequential decision making problem under the framework of RL. At each step, the agent receives an observation $o_t$ from the environment (i.e., the onboard sensor), decides and applies an action $a_t$ according to a policy $\pi$, and then observes a reward signal $r_t$. The goal of the agent is to learn an optimal policy that maximizes the expected sum of discounted rewards $R_t$ 
	\begin{equation}
	R_t = \sum_{i=t}^{T}\gamma^{i-t}r_i,
	\end{equation}
where $T$ is the terminated time step and $\gamma\in[0,1]$ is a discount factor that determines the importance of future rewards. The underlying state $s_t$ of the system includes the physical state configuration (positions, velocities, etc.) of both the quadrotor and the target object (which generally need to be inferred from the observations $o_t$). Since the actions $a_t$ consist of low-level motor commands, the complex system dynamics $p(s_{t+1}|s_t,a_t)$ make it difficult to learn a stable policy by directly applying existing model-free RL methods.

\subsection{Deep Deterministic Policy Gradient}
When the system dynamics $p(s_{t+1}|s_t,a_t)$ are not known, model-free RL methods such as policy gradient~\cite{peters2006policy} and Q-learning~\cite{sutton1999policy} methods are often preferred.
Assuming that the environment is fully observed so that $o_t=s_t$,
the Q-function $Q^\pi(s_t, a_t)$ represents the expected return after taking an action $a_t$ in state $s_t$ and thereafter following policy $\pi$:
\begin{equation}
Q^\pi(s_t,a_t) = \BE_\pi[R_t|s_t,a_t].
\end{equation}
Consider a Q-function approximator parameterized by $\theta^Q$.  It can be optimized by minimizing the loss:
\begin{equation}
L(\theta^Q) = \BE_\pi\left[\left(Q(s_t,a_t|\theta^Q)-y_t\right)^2\right],
\end{equation}
where $y_t = r(s_t,a_t)+\gamma \max_aQ(s_{t+1},a|\theta^{Q'})$ is the target Q-value estimated by a target Q-network.

However, for continuous action problems Q-learning becomes difficult since it requires maximizing a complex, nonlinear function at each update to improve the current policy. Therefore continuous domains are often tackled by actor-critic methods, where a separate parameterized ``actor" policy is learned in addition to the Q-function. The Deep Deterministic Policy Gradient (DDPG)~\cite{ddpg} algorithm, based on Deterministic Policy Gradient~\cite{dpg}, maintains a parameterized actor function $\mu(s|\theta^{\mu})$ which specifies the current policy by deterministically mapping each state to a unique action. The actor is updated by performing gradient ascent based on the following policy gradient:
\begin{equation}
\nabla_{\theta^{\mu}}\mu \approx
\BE_{\mu'}[\nabla_a Q(s,a|\theta^Q)|_{s=s_t,a=\mu(s_t)} \nabla_{\theta^{\mu}}\mu(s|\theta^{\mu})|_{s=s_t} ].
\end{equation}
By incorporating the ideas of sample replay buffer and target network backup originated from DQN~\cite{dqn}, DDPG can use neural network function approximators for problems that involve continuous action domains.

\section{Our Approach}
Although model-free methods such as DDPG allow us to optimize complex policies based on raw image observations, they require massive amounts of data to achieve good performance. Besides, function approximators such as neural networks defined on high-dimensional observation spaces are very difficult to train in fragile physical systems such as quadrotors, since the agent can hardly find actions to reach any good state in the exploration process, especially in (infinite) continuous action domains. We now present our proposed approach which is particularly suitable for this task.
\begin{figure*}[htb]
	\centering
	\includegraphics[width=0.8\textwidth]{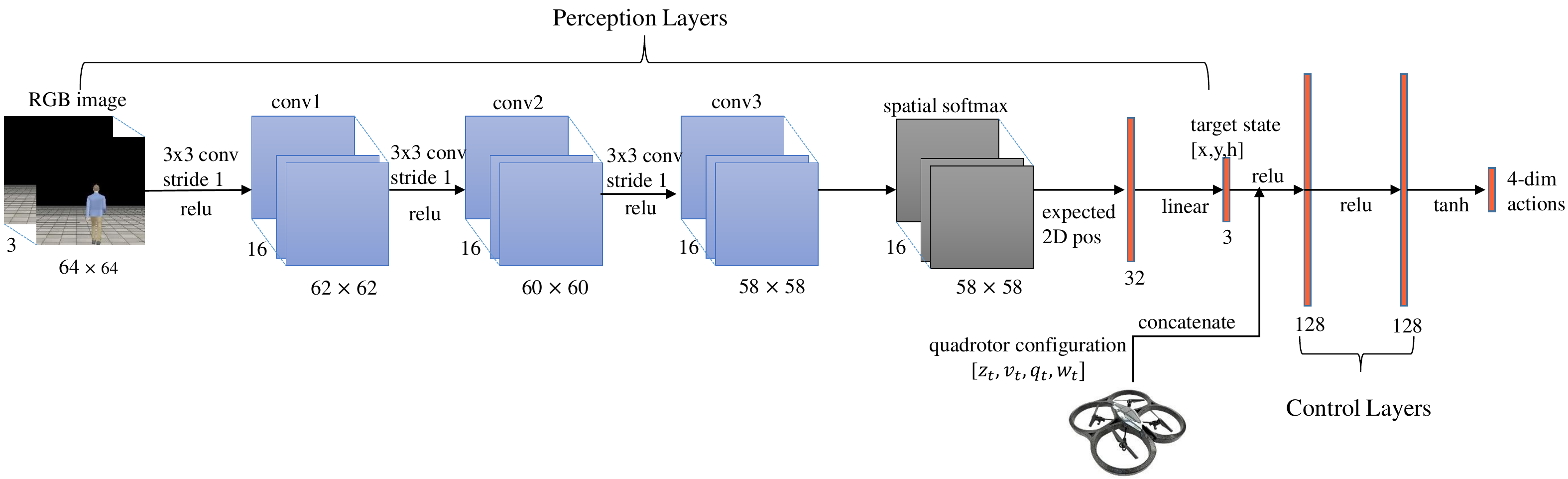}
	\caption{\label{fig:archi}
		Policy network architecture.
	}
\end{figure*}
\subsection{Policy Network Architecture}
In order to represent a policy that performs both perception and control, we use deep neural networks. As shown in Figure~\ref{fig:archi}, the policy network maps monocular RGB images and quadrotor configurations to the actions. 
The visual processing layers have minor differences from the conventional network architectures used for image classification. Pooling is discarded due to the loss of spatial information. 
Inspired by~\cite{deep-visuo}, we incorporate a spatial-argmax layer after the last convolutional layer to convert each pixel-wise feature map into spatial coordinates in the image space.
The spatial-argmax layer consists of a spatial softmax function applied to the last convolutional feature map and a fixed sparse fully connected layer which calculates the expected image position of each feature map. 
The spatial feature points are then regressed to a three-dimensional vector, $\s_{o,t}=(x_t, y_t, h_t)$, which represents the 2D position and scale (here we only keep the height information) of the target on the image plane, by another fully connected layer.
In order to achieve stable flight, it is essential to use the quadrotor configuration $\s_{q,t} = (z_t, \v_t, \q_t, \w_t)$, which includes the altitude, linear velocity, orientation, and angular velocity, as an additional input to the neural network.
After the visual processing layers, the target related state $\s_{o,t}$ is concatenated with the quadrotor state $\s_{q,t}$, followed by fully connected layers to the actions.

While in principle we could choose to directly output low-level motor actions $a_t$ by the policy network, the agent will be stuck to yield little performance improvement even after tens of thousands of sample experiences. We reason that pure model-free RL methods cannot effectively learn stable policies in such fragile systems. Therefore we introduce another set of high-level actions $u_t$ as the output of the policy network. The high-level actions are then mapped to low-level motor commands by a PID controller.

\begin{figure}[htb]
	\centering
	\includegraphics[width=0.45\textwidth]{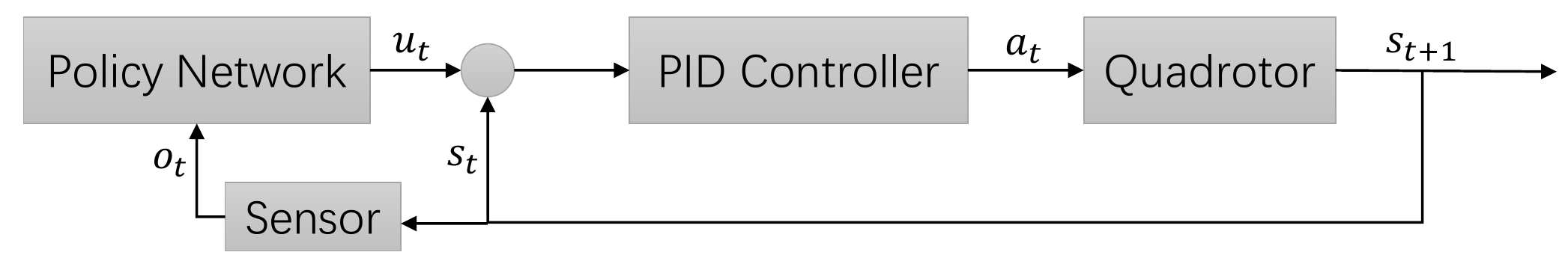}
	\caption{\label{fig:pid}Hierarchical control system with the policy network.}
\end{figure}

\subsection{Combination with PID Control Loop}
\label{sec:pid}
The quadrotor features highly nonlinear dynamics and complex aerodynamics which are hard to learn by model-free RL methods. Fortunately, this challenge can be tackled by incorporating a conventional PID controller.
Figure~\ref{fig:pid} shows the proposed hierarchical control system. At each time step $t$, given the observed image, the policy network generates a four-dimensional high-level reference action $u_t$ 
\begin{equation}
u_t = (p_x,p_y,p_z,\varphi_{yaw}),
\end{equation} 
which corresponds to the relative position offset in the $x$-, $y$-, $z$-directions and the relative angle offset around the yaw axis.
Thanks to the differential flatness property proposed in~\cite{minimum_snap}, the above high-level reference trajectories can be followed by simple cascade PID controllers with dynamical feasibility guarantee.
With the help of the PID controller, the agent can now learn the complex behavior in a more stable and effective manner. 
While the quadrotor dynamics of simulated models are often significantly different from those of real-world platforms, their high-level decision strategies are generally very similar. Thus introducing high-level actions also makes it much easier to transfer a policy learned in a simulator domain to a real-world domain.

\subsection{Reward Shaping}
Due to the continuous control nature of the target following task, an immediate reward feedback at every time step is essential. Besides, the instability and fragility of the quadrotor system also poses some additional challenges to the design of the reward function. A naive reward function based solely on the target related state $\s_{o,t}$ will lead to suboptimal policies that cannot guarantee flying stability.  Therefore the reward function should consider both the quadrotor state and the target related state.

To that end, we design the reward function as a combination of the goal-oriented target reward and the auxiliary quadrotor reward:
\begin{equation}
r = r_o(\s_o) + r_q(\s_q).
\end{equation}
Note that for notational simplicity we omit the time step $t$ from the subscript here.
By overloading the symbol $r_o$, the target reward is expressed as the sum of two parts as below:
\begin{equation}
r_o(\s_o) = r_o(x,y) + r_o(h),
\end{equation}
which correspond to the position reward and scale reward, respectively.
Let $s_{part}$ denote part of the state $\s_o$, which corresponds to either $(x,y)$ or $h$. Then the corresponding reward takes the following form:
\begin{equation}
\label{eqn:reward_goal}
r_o(s_{part}) = \left\{
\begin{array}{lr}
\exp(-\Delta s_{part}) & \Delta s_{part} \leq \tau_1 \\
0 & \tau_1 < \Delta s_{part} \leq \tau_2 \\
-(\Delta s_{part} - \tau_2) & \Delta s_{part} > \tau_2
\end{array}
\right.,
\end{equation}
where $\Delta s_{part}= \|s_{part} - s^*_{part}\|_2$ denotes the $\ell_2$-norm  between the current state and the desired goal state. 
The intuition is that the learner must observe variations in the reward signal in order to improve the policy. This hierarchical form essentially provides an intermediate goal to guide the learning process to find a reasonable solution step by step.
Also with a slight abuse of notation, the auxiliary reward is expressed as follows:
\begin{equation}
r_q(\s_q) = r_q(z) + r_q(q_1,q_2,q_3,q_4),
\end{equation}
which correspond to the quadrotor altitude and orientations (expressed as quaternions), respectively.
Different from the target reward, the auxiliary reward is used to impose additional constraints on the flying gesture. Only penalty terms are introduced.
By using the same notations as in the target reward, the auxiliary reward takes the following form:
\begin{equation}
r_q(s_{part}) = \left\{
\begin{array}{lr}
{-c}(1 - \exp(-\Delta s_{part}))& \Delta s_{part} \leq \tau_1 \\
{-c} & \Delta s_{part} > \tau_1
\end{array}
\right.,
\end{equation}
where $\tau_1$ denotes the same threshold as in equation (\ref{eqn:reward_goal}) and $c$ denotes the penalty weight.

With the reward function defined above, existing policy gradient methods (such as DDPG) can be applied in a game playing environment to train the policy network.

\subsection{Training Strategy}
\label{sec:pretrain}
Although in principle the policy network can be directly trained end-to-end by DDPG, our empirical finding shows that the agent will suffer severe vibration in the learning process, mainly due to the high sample complexity of model-free algorithms. Another drawback is that the perception layers cannot be guaranteed to accurately locate the target of interest.

We propose to introduce a supervised pre-training stage that allows us to initialize the perception layers of the policy network using a relatively small number of training iterations.
The dataset can be easily collected by randomly moving the quadrotor and recording the camera image stream. In a simulator, we can directly get the noiseless labels. In real-world domains, we can use existing model-free object trackers~\cite{drone-tracking} to get the noisy labels. Both are sufficient to train a pose regression CNN. After training on the regression task, the weights of the visual processing layers can be transferred to the policy network. 
By factoring the perception and control tasks in learning, we can gain the image generalization power of CNNs across different environments (simulator and real-world). This is more reasonable than some existing approaches~\cite{caduav} which directly transfer simulator perception to the real world.

After the supervised learning stage, we first fix the perception layers and learn only the weights of the fully connected control layers which are not initialized with pre-training. Then the entire policy network is further optimized in an end-to-end manner. Empirical findings show that jointly optimizing the whole network from the very beginning hurts the pre-trained representation ability of the perception layers.



\subsection{Transfer to the Real World}
As explained in Section~\ref{sec:pid}, learning the high-level actions makes it easier to transfer a policy from the simulator domain to a real-world domain. Considering the large gap between the simulated images and the real-world images, we take the factorizing scheme during training as in \mbox{Section~\ref{sec:pretrain}}. Namely, the perception layers are pre-trained by a small dataset collected in real-world scenarios. There are also several challenges to set up a game playing environment for real-world quadrotors. First, we cannot access the true target state efficiently online and thus cannot reliably compute the reward. Second, efficiently resetting the game state upon game termination is difficult. Fortunately we can bypass these issues to directly use the policy behavior trained in simulators. Since in simulators all the underlying true states are available, we can directly learn the control layers with the state input. Finally the perception layers and the control layers are merged to form the policy network which is applicable to real-world domains.

\section{Experiments}
In this section we present a series of experiments to answer the following research questions\footnote{You can view a movie of the learned policies at \url{https://youtu.be/cooKHBtIpr0}}:
\begin{compactenum}
	\item Is introducing the PID controller essential for successful training?
	\item How does the training strategy work compared to standard end-to-end training?
	\item How does the learned high-level policy network generalize across different environments?
\end{compactenum}
To answer question (1) and (2),	we evaluate different variations of the proposed system, in Section~\ref{sec:system_design_evaluate}, by training policies for the target following task in a simulated environment.
This allows us to validate the importance of each individual component in the proposed approach.
We further evaluate the generalization ability of the learned policy in Section~\ref{sec:policy_eval} by testing it in various simulated environments.
Finally, we set up a real-world flight test in Section~\ref{sec:policy_transfer} by directly transferring our approach from the simulator to a DJI quadrotor platform.

\subsection{Simulator Settings}
\begin{figure*}[htb]
	\centering
	\subfigure[Simulated scenario 1.]{\includegraphics[width=0.24\textwidth]{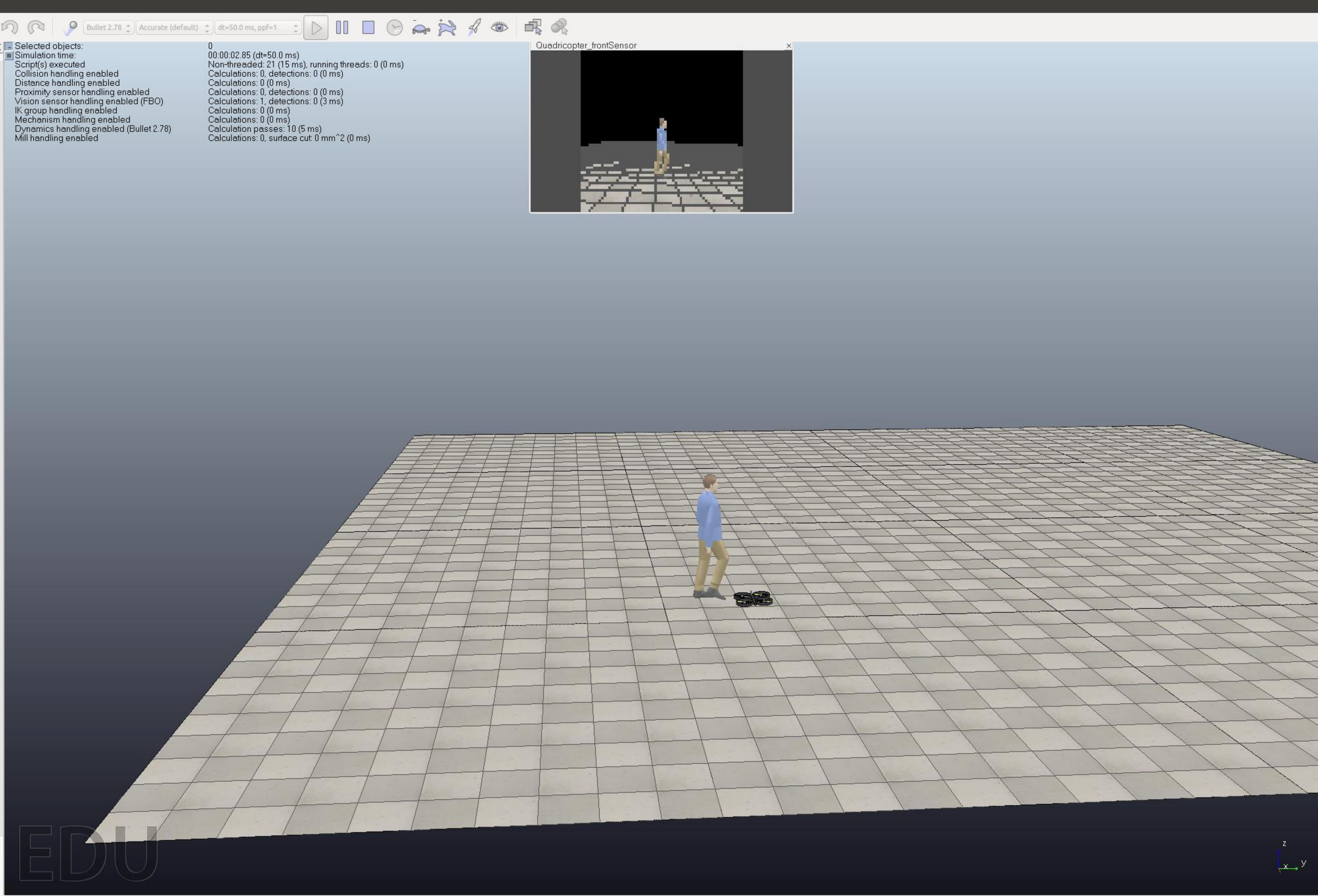}}
	\subfigure[Simulated scenario 2.]{\includegraphics[width=0.24\textwidth]{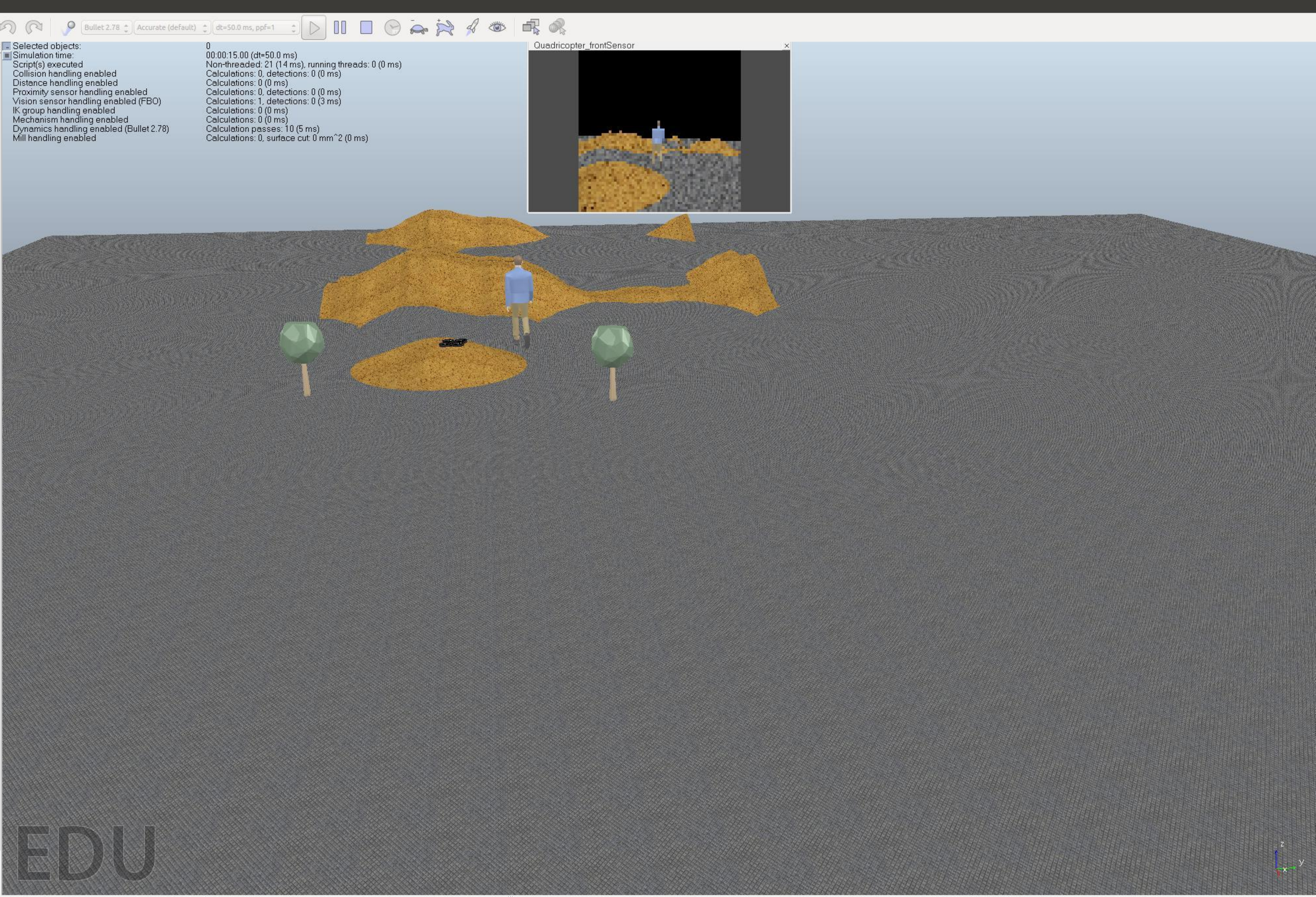}}
	\subfigure[Simulated scenario 3.]{\includegraphics[width=0.24\textwidth]{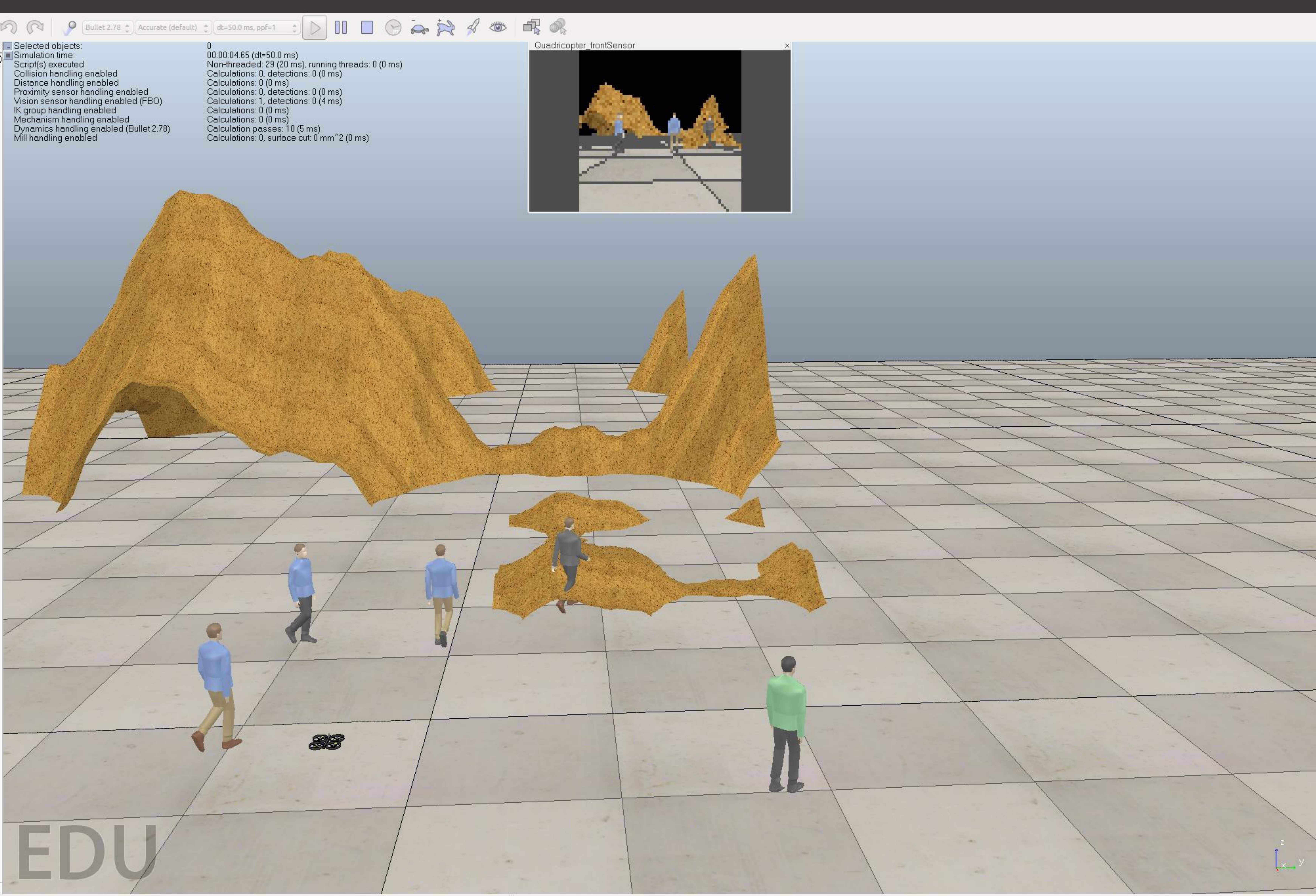}}
	\subfigure[Real-world scenario.]{\includegraphics[width=0.24\textwidth]{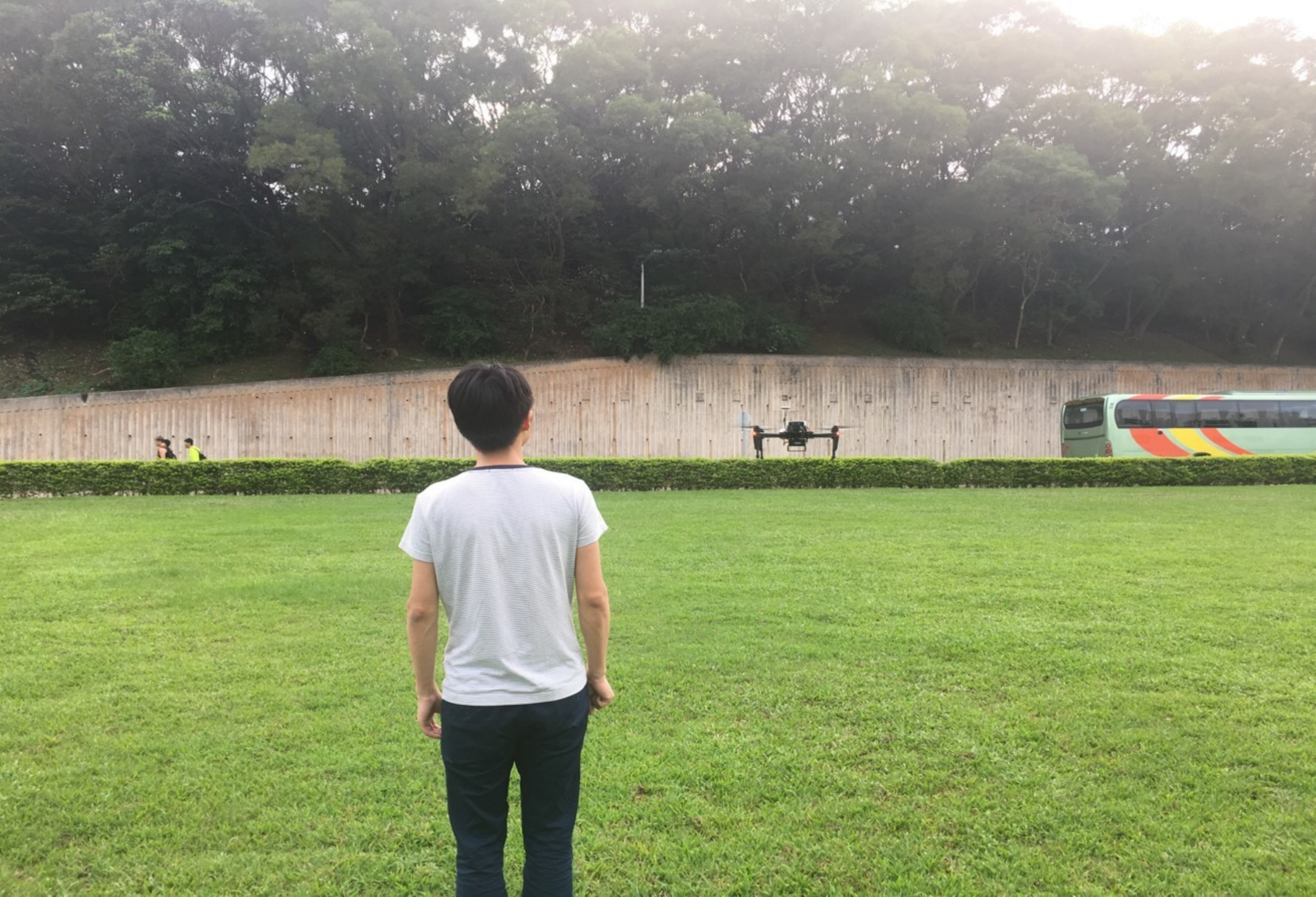}}
	\caption{\label{fig:target_task}Three simulated environments and one real-world environment.}
\end{figure*}

\textbf{Environment.}
We set up the simulated target following task on the Virtual Robot Experimentation Platform (V-REP) using the built-in quadrotor model. The observed state of the quadrotor $\s_{q,t} = (z_t, \v_t, \q_t, \w_t) \in \BR^{11}$ consists of the altitude, linear velocity, orientation, and angular velocity, where the velocities are expressed in the body frame.
The state of the target object is unknown and must be inferred from the RGB image input with resolution $64\times 64$.
We require the target to be in the camera view of the quadrotor on initialization.  At each time step, the target randomly chooses a direction to walk at a random speed. The game will terminate either when the target is out of the camera view or when the quadrotor crashes, as determined by using simple thresholds on the quadrotor's altitude and orientation. The maximum episode length allowed is set to 1000. Three different simulated scenes are shown in the left part of Figure~\ref{fig:target_task}.

\textbf{Implementation Details.}
We choose the off-policy actor-critic algorithm DDPG to train the policy network due to its sample efficiency over on-policy policy gradient methods.
Our implementation is based on \mbox{rllab}~\cite{rllab}.
The Q-network shares the same architecture with the policy network, except that the last two fully connected layers have a smaller number of units (32) and the actions are included in the second to last layer. We use Adam~\cite{adam} for optimization with the hyperparameters set according to~\cite{ddpg}. For the reward setting, we use $\tau_1=0.05$, $\tau_2=0.2$, and $c=0.5$.

\subsection{System Design Evaluation}
\label{sec:system_design_evaluate}

\begin{figure*}[htb]
	\centering
	\includegraphics[width=0.24\textwidth]{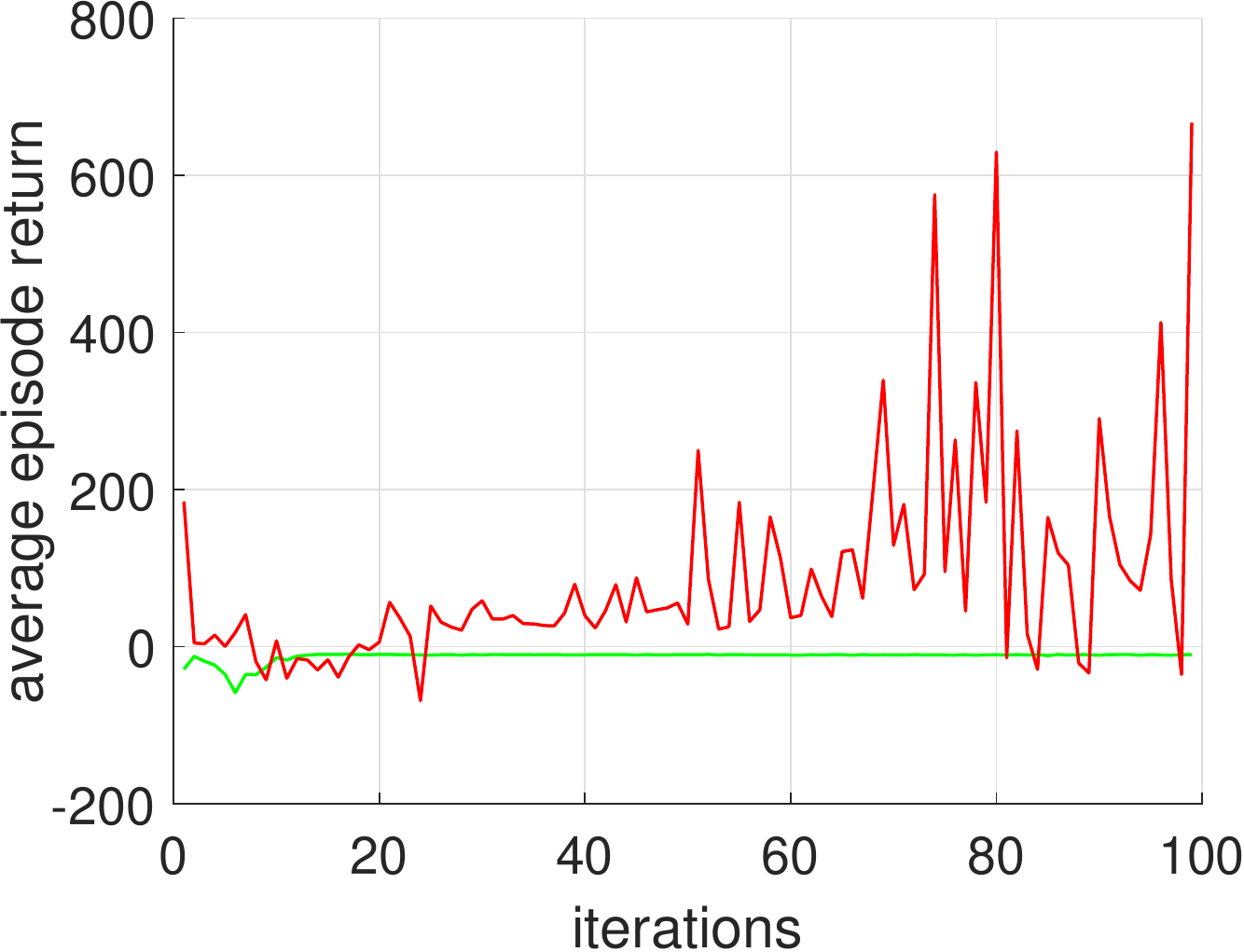}
	\includegraphics[width=0.24\textwidth]{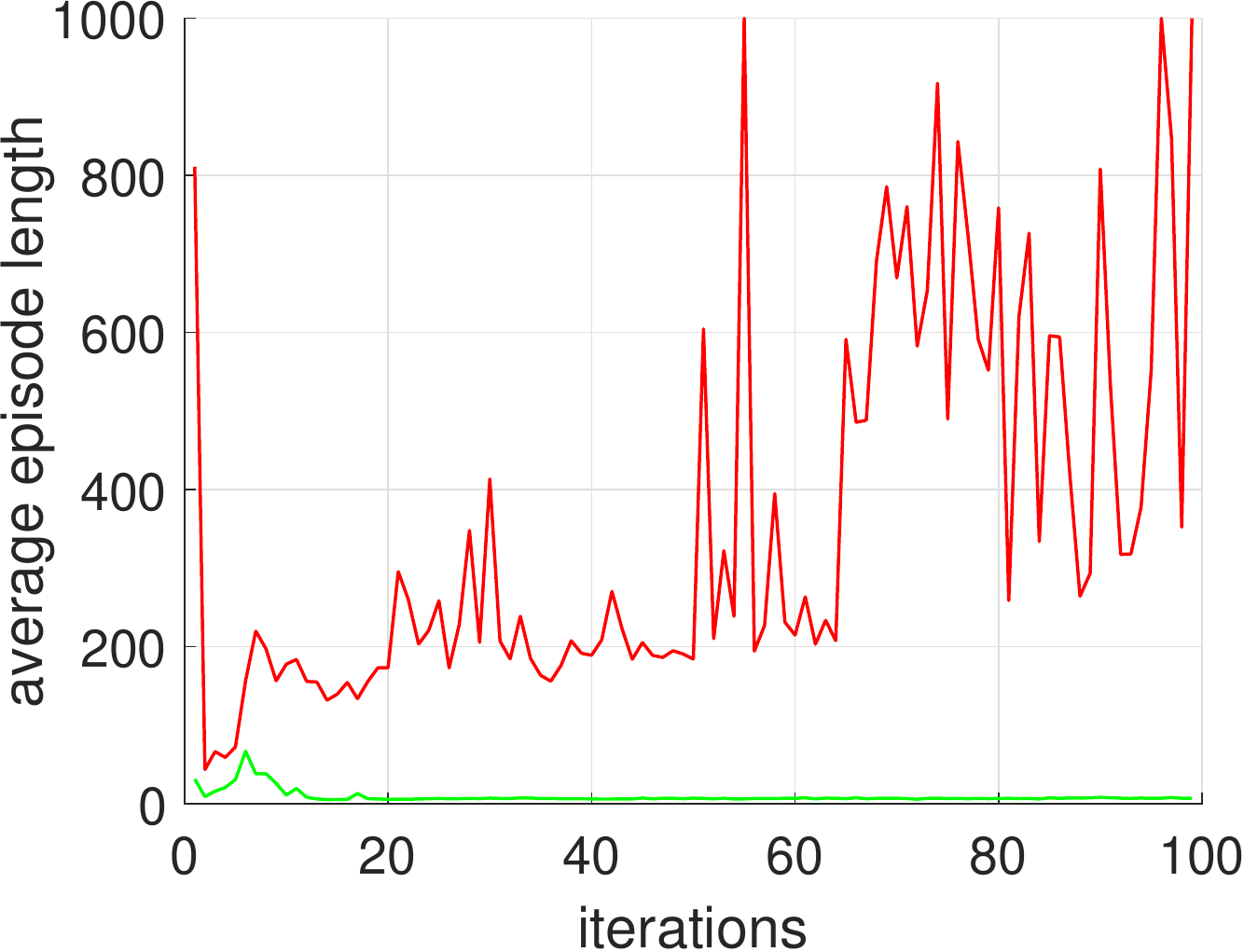}
	\includegraphics[width=0.24\textwidth]{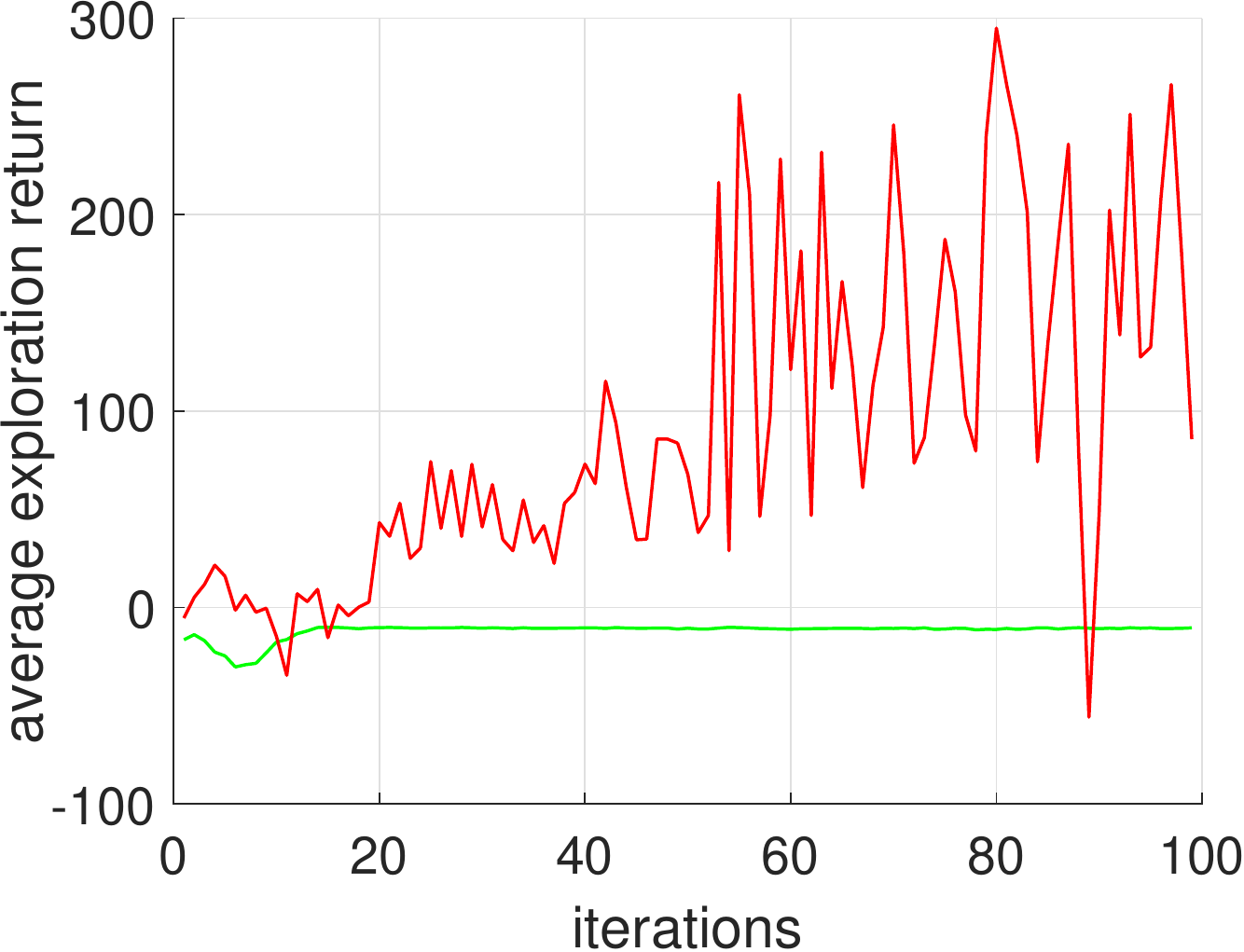}
	\includegraphics[width=0.24\textwidth]{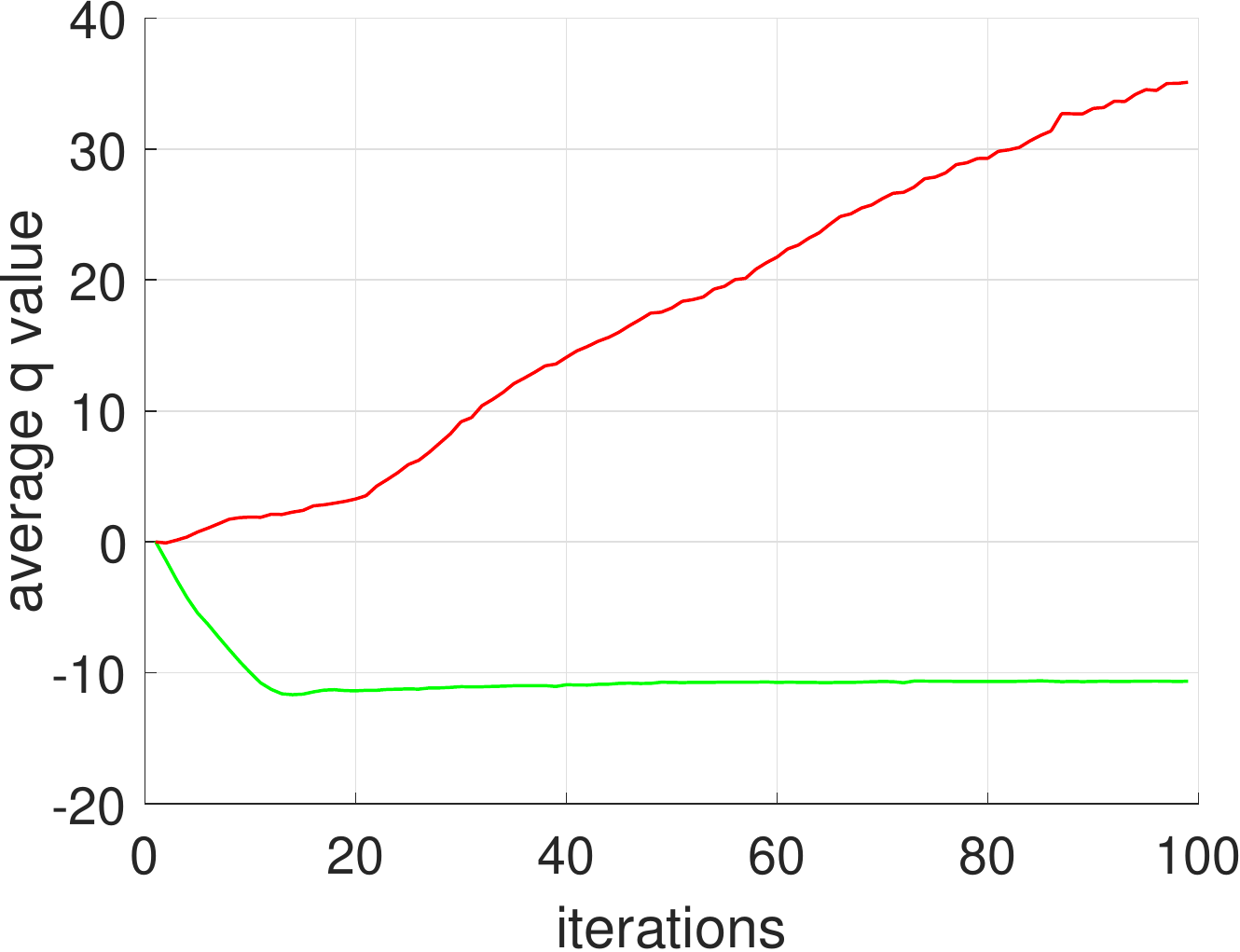}\\
	\includegraphics[width=0.4\textwidth]{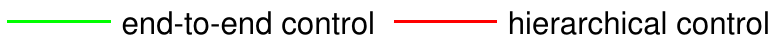}
	\caption{\label{fig:pid_comparison}Training results of the end-to-end motor-level control system and the proposed hierarchical control system.
		The hierarchical control system is substantially easier to train.
	}
\end{figure*}
\begin{figure*}[htb]
	\centering
	\includegraphics[width=0.24\textwidth]{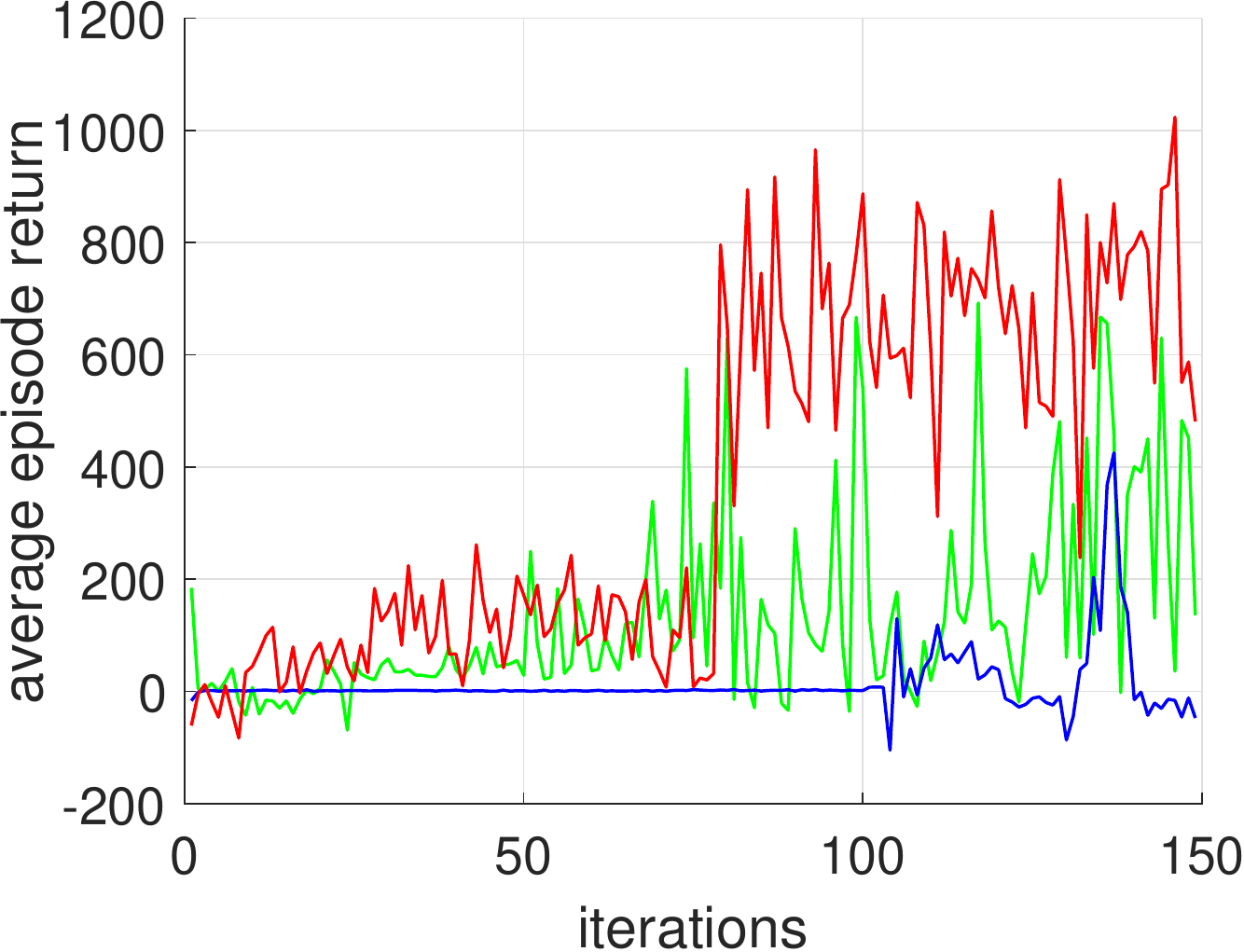}
	\includegraphics[width=0.24\textwidth]{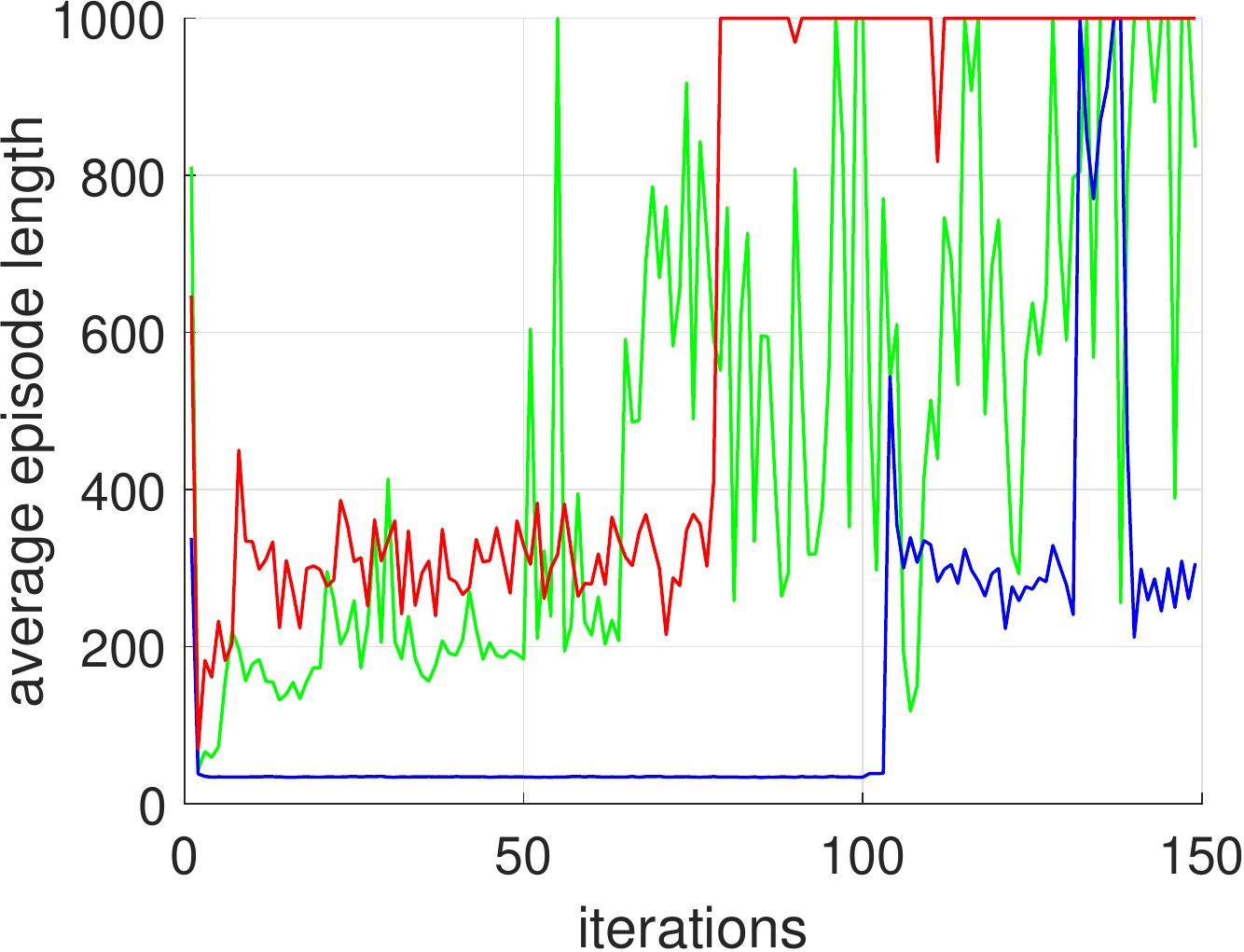}
	\includegraphics[width=0.24\textwidth]{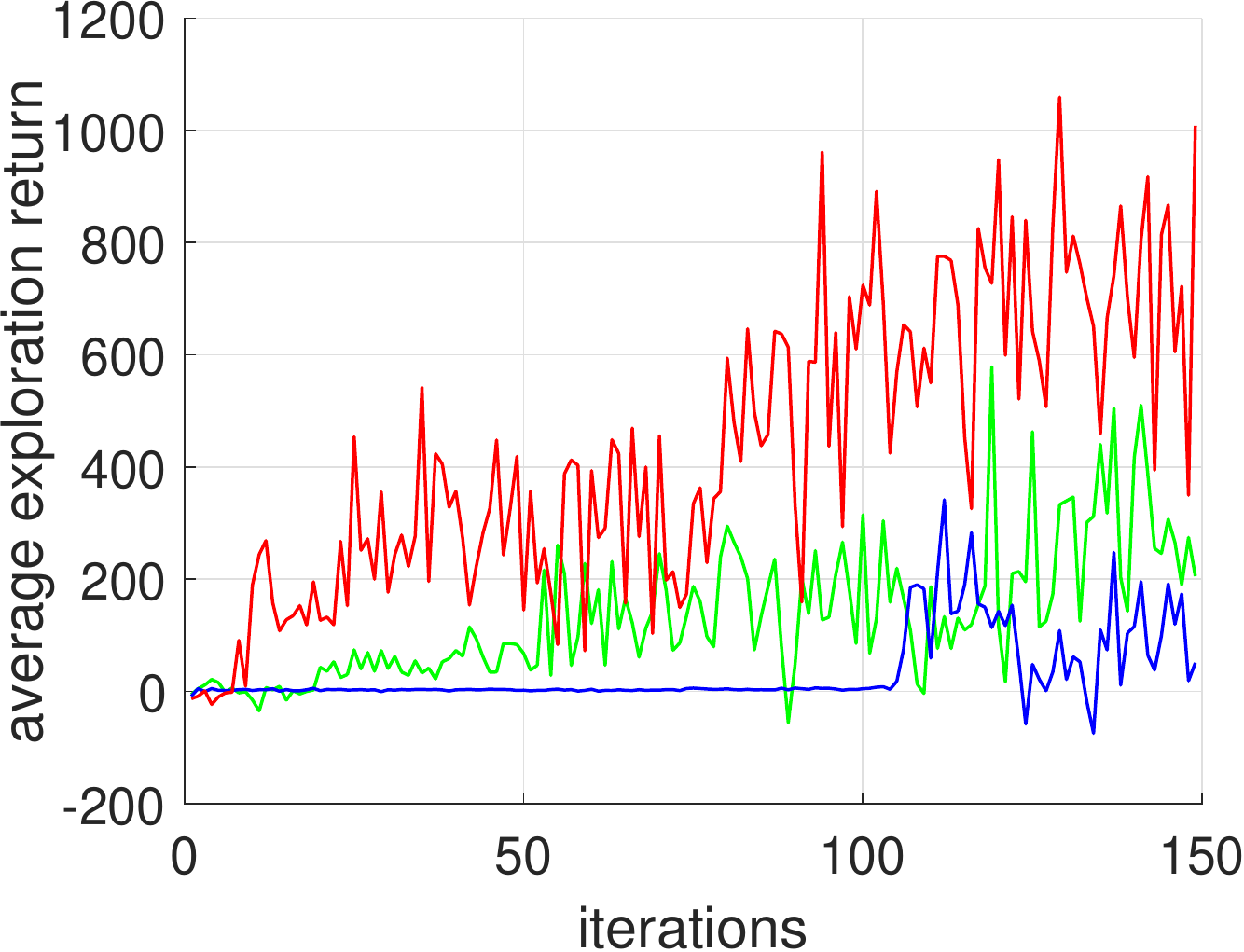}
	\includegraphics[width=0.24\textwidth]{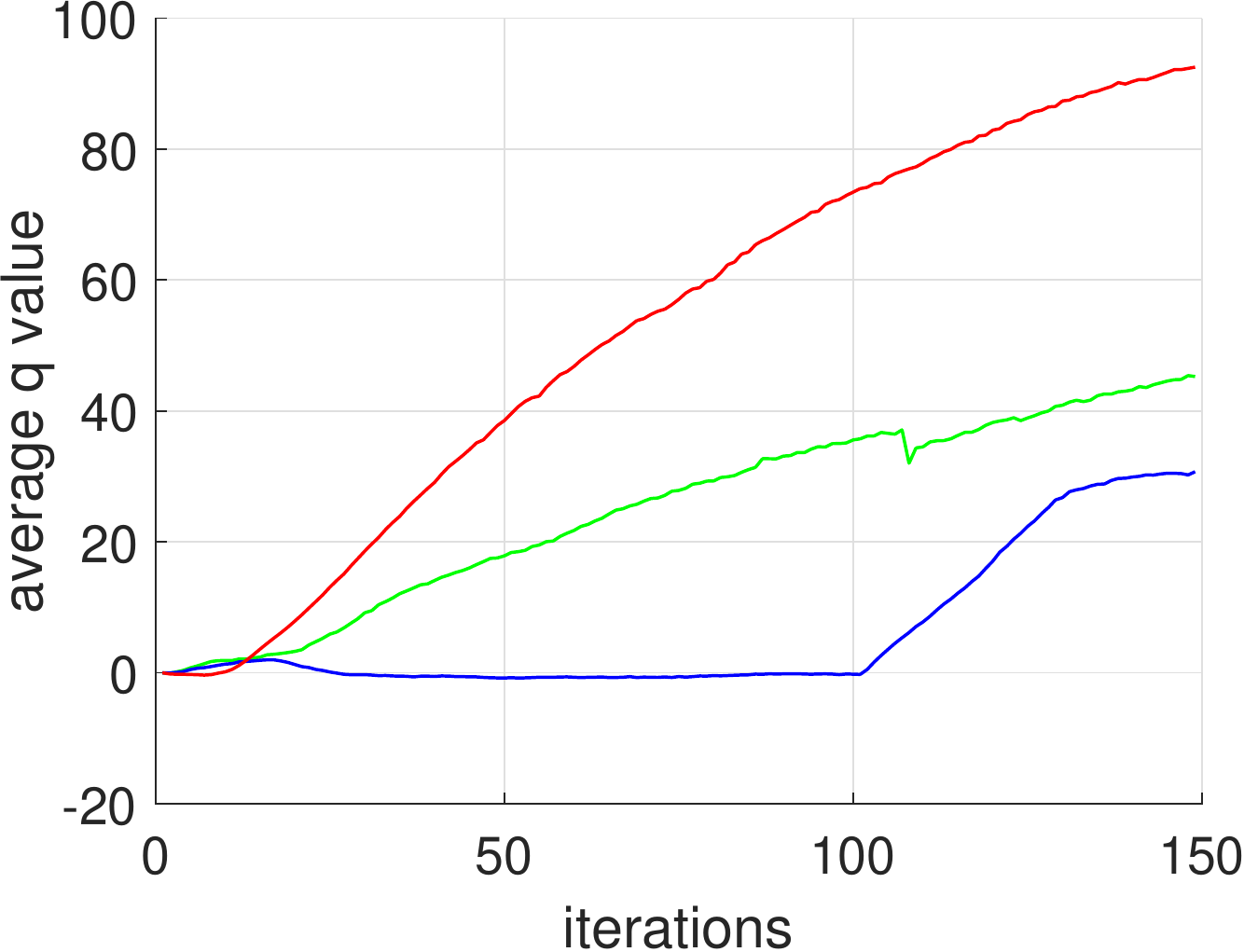}\\
	\includegraphics[width=0.5\textwidth]{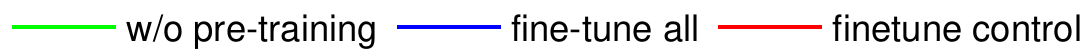}
	\caption{\label{fig:3_comparison}Training results on three different approaches. 
	}
\end{figure*}

\textbf{Hierarchical control system vs.\ end-to-end control system}
In this experiment, we show the effectiveness of introducing the PID controller to the control system. To that end, we compare two approaches: the first one being an end-to-end control system where the policy network directly outputs $a_t$ and the second one being the hierarchical control system with PID where the policy network outputs $u_t$. Both approaches are trained with standard training strategy (no pre-training and no hierarchical fine-tuning). Figure~\ref{fig:pid_comparison} shows the learning curves of these two different approaches. Intuitively, the average return measures the precision of control, while the the average episode length measures the successfully followed steps (robustness). We also show the average exploration return curve and estimated Q-value curve.
In both methods, the network is trained for 100 iterations with 2000 time steps for each iteration. We can see that a direct application of DDPG to learn low-level motor commands cannot achieve any performance improvement while the proposed hierarchical control approach achieves substantial improvement as learning proceeds.

\begin{figure*}[htb]
	\centering
	\includegraphics[width=0.24\textwidth]{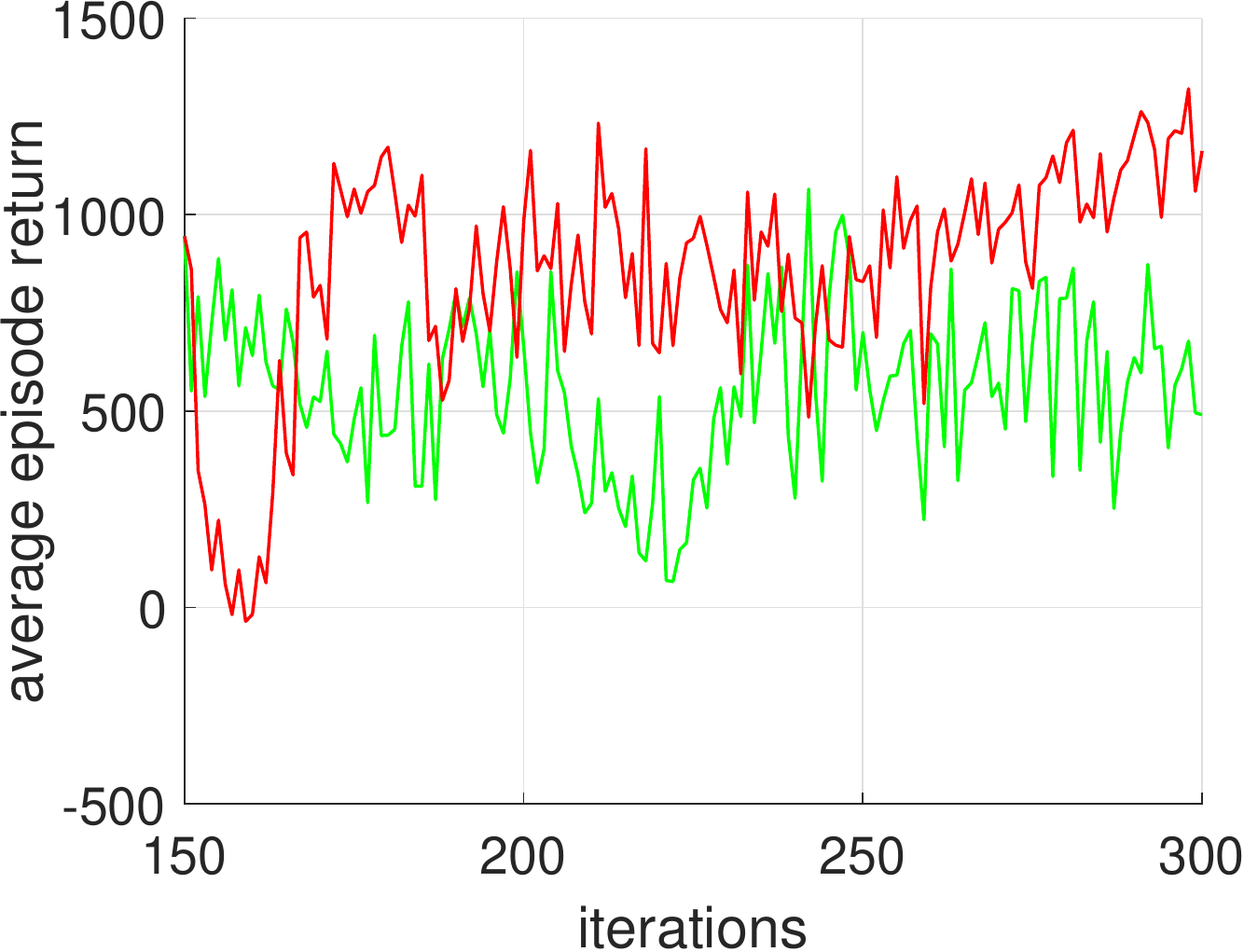}
	\includegraphics[width=0.24\textwidth]{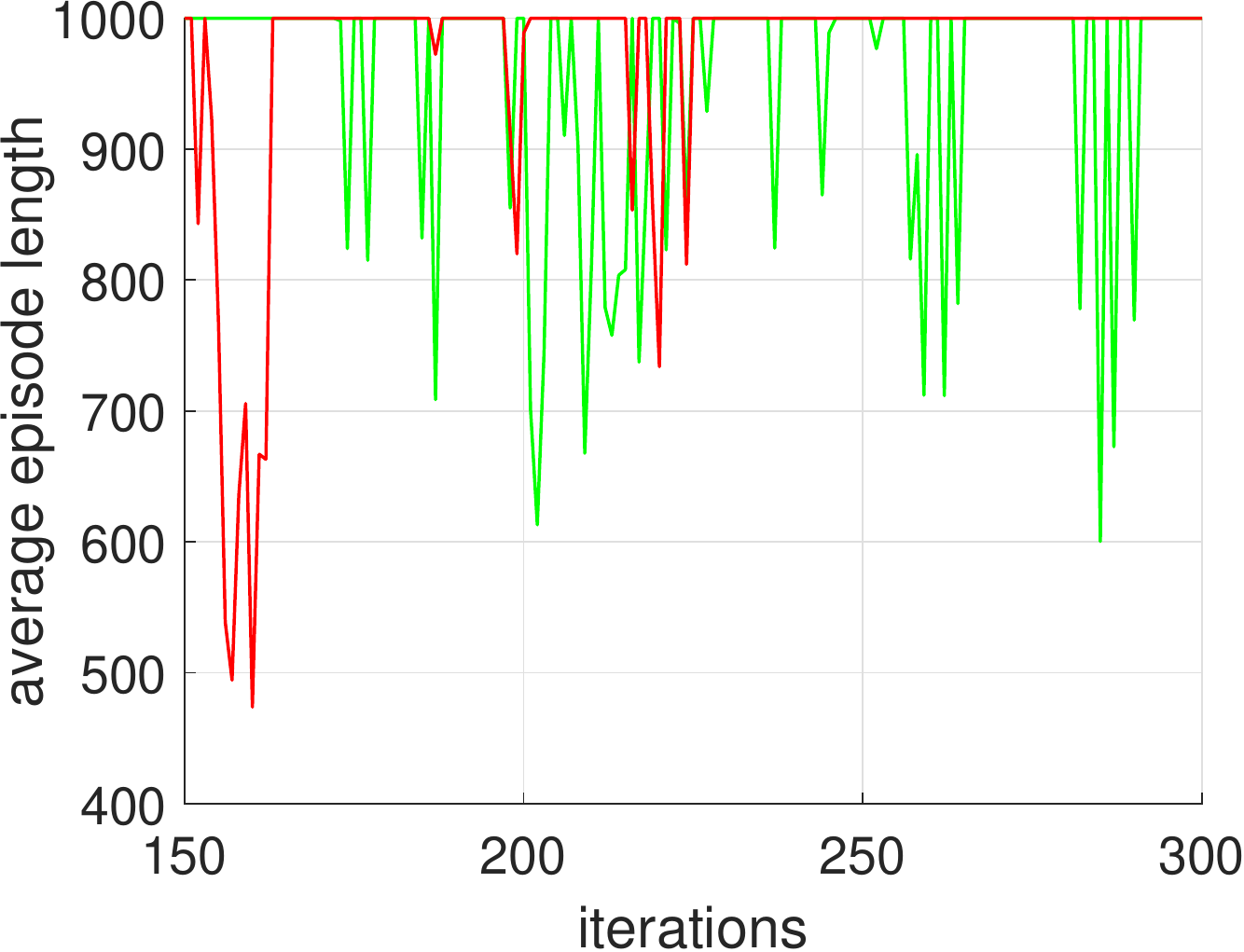}
	\includegraphics[width=0.24\textwidth]{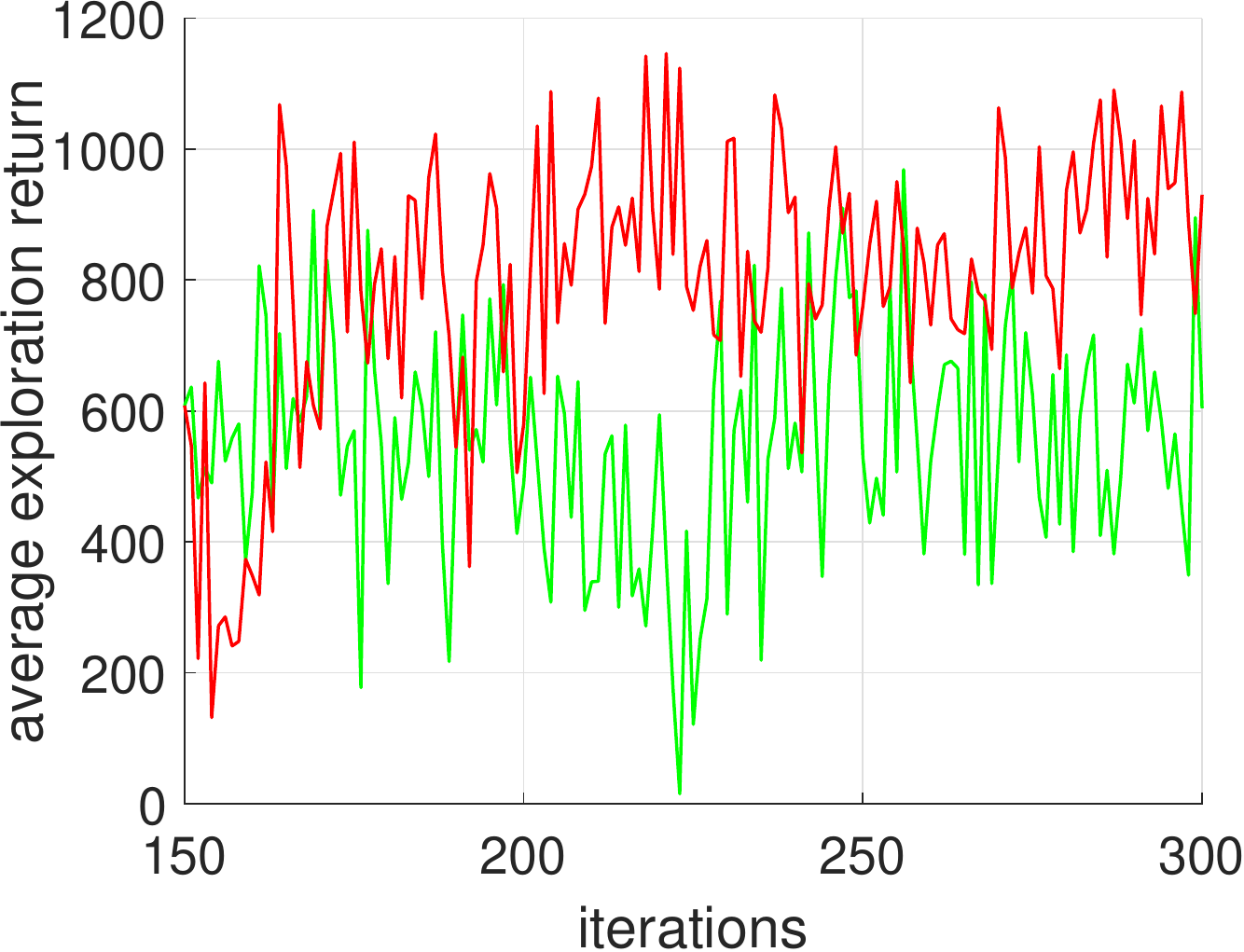}
	\includegraphics[width=0.24\textwidth]{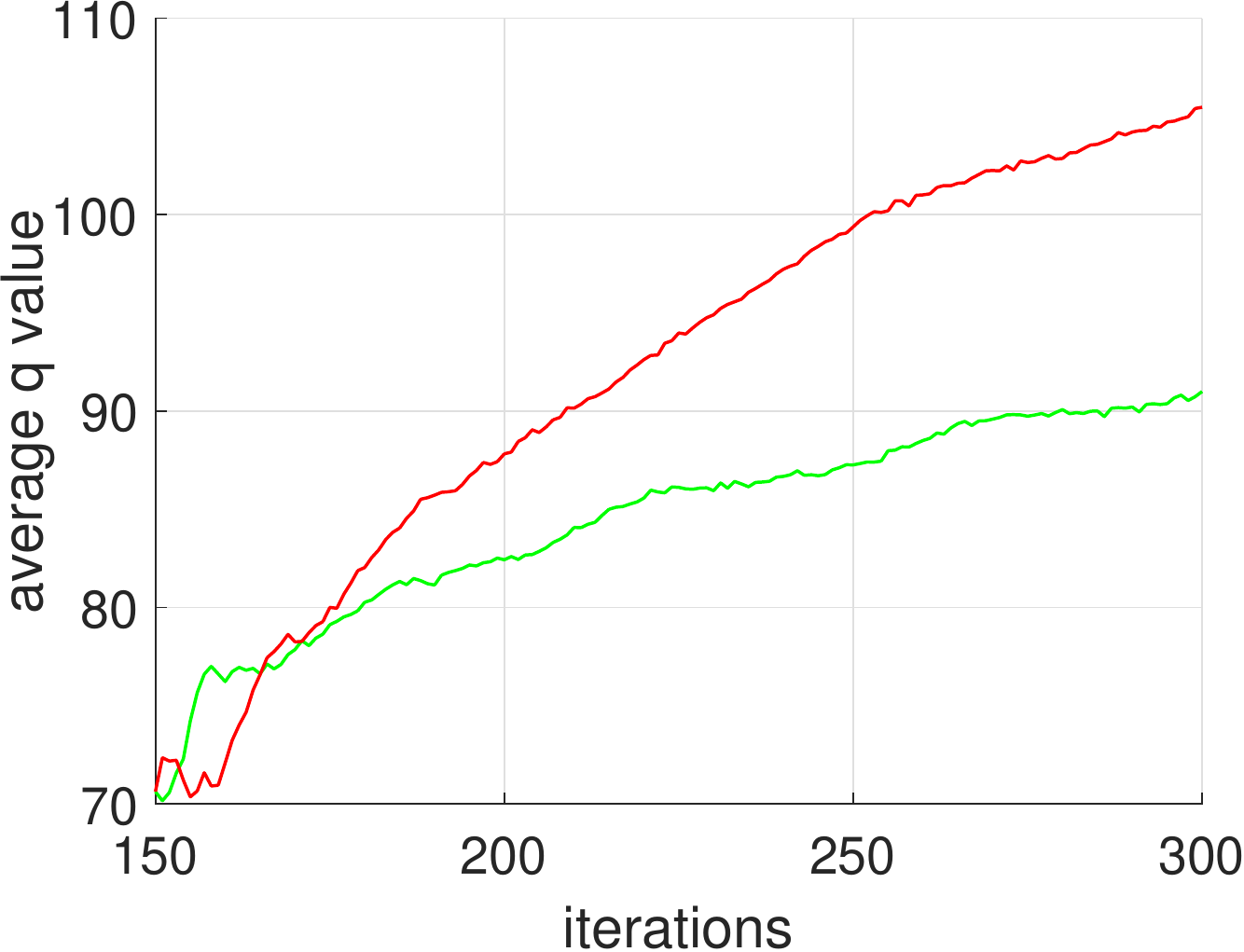}\\
	\includegraphics[width=0.4\textwidth]{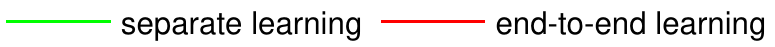}
	\caption{\label{fig:2_comparison}Training results between the separate learning approach and the end-to-end learning approach. In both methods, the first 150 iterations are trained by only fine-tuning the control layers in Figure~\ref{fig:3_comparison}. 
	}
\end{figure*}

\textbf{With pre-training vs.\ without pre-training}
We can observe in Figure~\ref{fig:pid_comparison} that the agent suffers severe vibration in the learning process under standard training strategy.
To validate the necessity of supervised pre-training, we compare three different approaches here. The first one is the standard training without any pre-training (\textbf{w/o pre-training}). The second one is to fine-tune all layers after pre-training (\textbf{fine-tune all}). The last one is our strategy which only fine-tunes the control layers after the pre-training initialization (\textbf{fine-tune control}).
Figure~\ref{fig:3_comparison} shows the performance of different methods.
We show the learning curves for up to 150 iterations since we observe only minor improvements afterwards. 
The results demonstrate that supervised pre-training of the perception module greatly increases the stability of the policy network, which has always been a major drawback of many RL algorithms. Without pre-training, the policy suffers severe vibration.
Actually, further experiments show that even with more training iterations, the pure DDPG algorithm still gets stuck in suboptimal policies. Another important observation is that jointly optimizing the whole network from the very beginning
might hurt the overall performance since it leads the convolutional layers to forget useful features learned through pre-training.

\textbf{End-to-end learning vs.\ separate learning}
So far, the perception layers and the control layers have been learned separately. We now examine our design choice of end-to-end fine-tuning: does training the perception and control layers jointly  provide better performance? After initializing the control layers as above, we fine-tune the whole network in an end-to-end manner. 
We also make comparison with a baseline in which only the control layers are fine-tuned.
Figure~\ref{fig:2_comparison} compares the separate learning approach and the end-to-end learning approach. The learning curve suggests that jointly training the perception and control layers end-to-end does further boost the performance.

\subsection{Policy Evaluation in Simulators}
\label{sec:policy_eval}
To gain more insights into how the learned policy actually works, we further apply the trained policy network in a number of simulated testing environments in which the agent interacts with the environment until game termination. For testing environment initialization, we randomly set the positions of the quadrotor and the target, making the target appear at different corners of the camera view with different scales.
The performance of the policy is measured by the deviation of the target state from our desired goal (which is specified in the reward function).

\begin{figure*}[htb]
	\centering
	\includegraphics[width=0.6\textwidth]{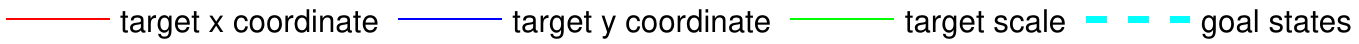}\\
	\subfigure[\label{fig:test_state}Separate vs.\ end-to-end in scenario~1 (left to right).]{\includegraphics[width=0.24\textwidth]{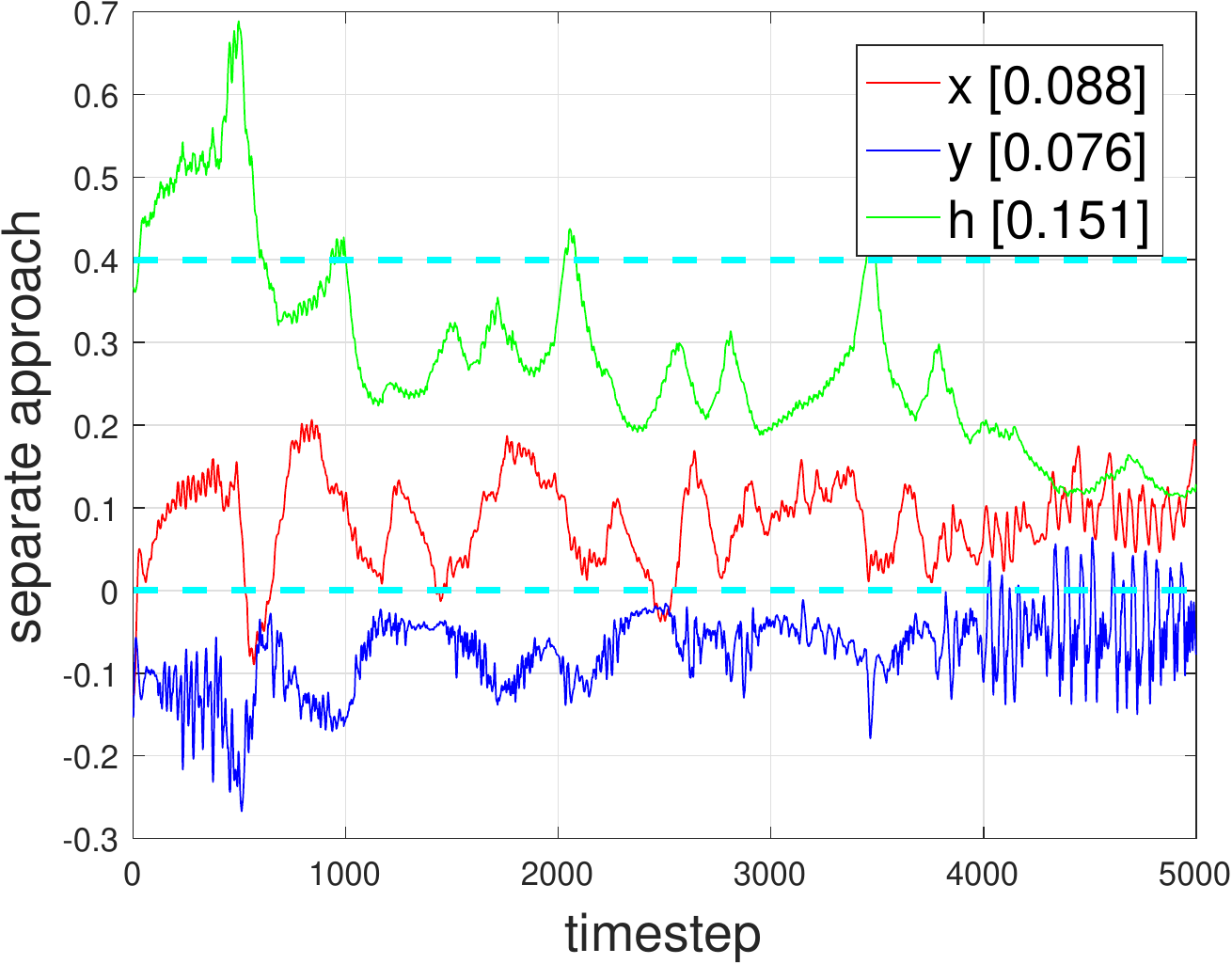}
		\includegraphics[width=0.24\textwidth]{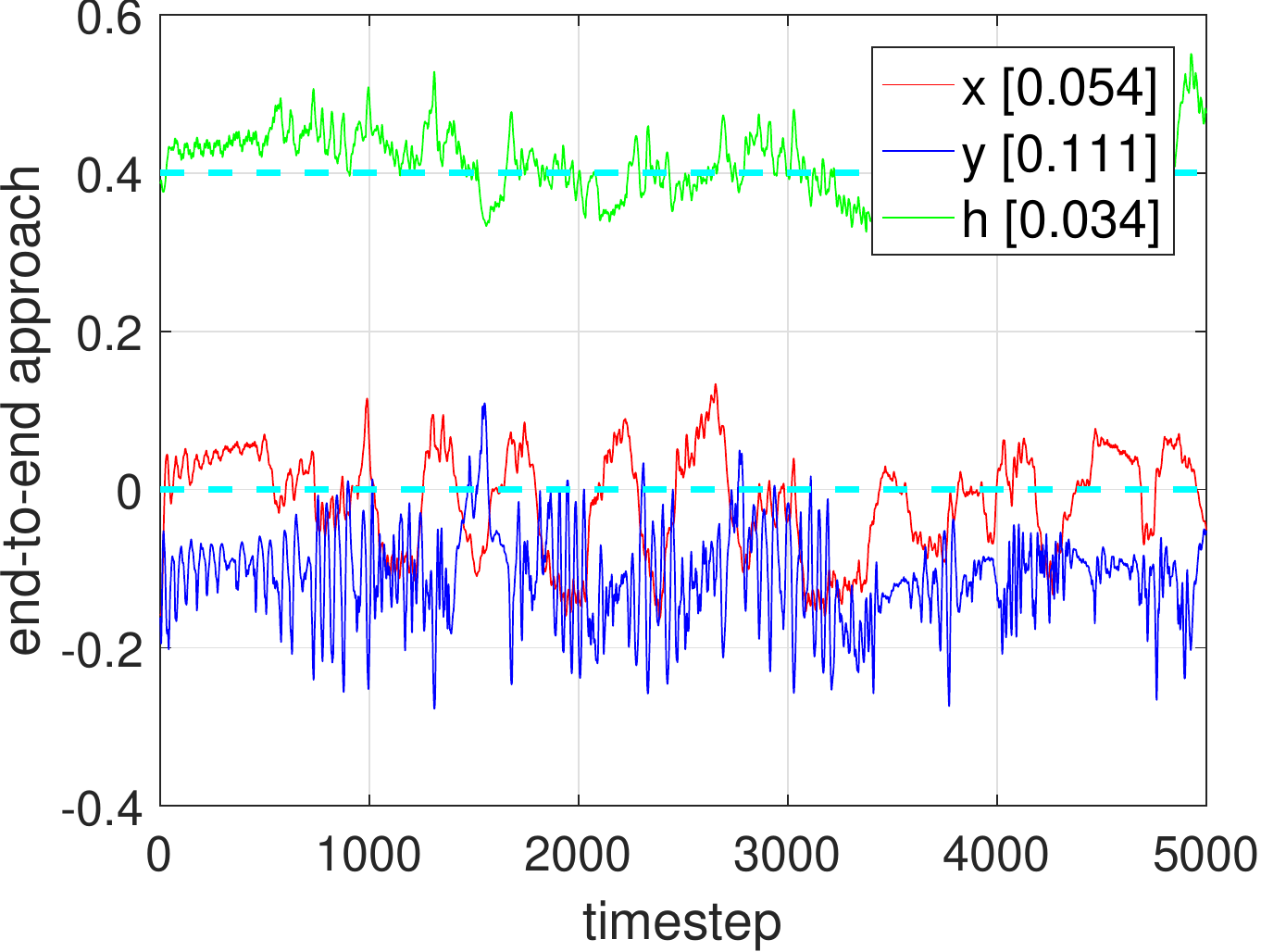}}
	\subfigure[\label{fig:test_state_unseen}Unseen scenarios~2 and~3 (left to right).]{\includegraphics[width=0.24\textwidth]{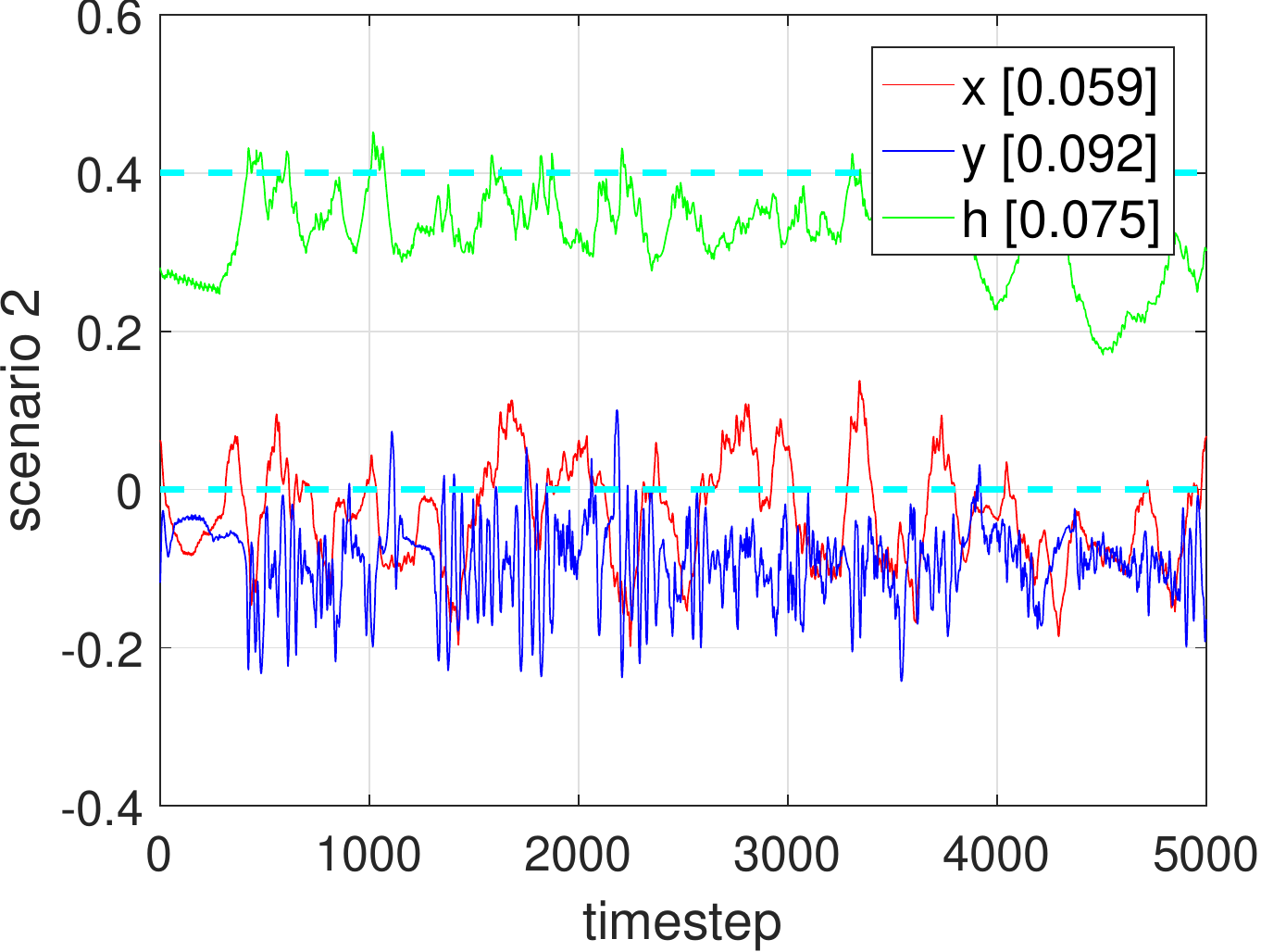}
		\includegraphics[width=0.24\textwidth]{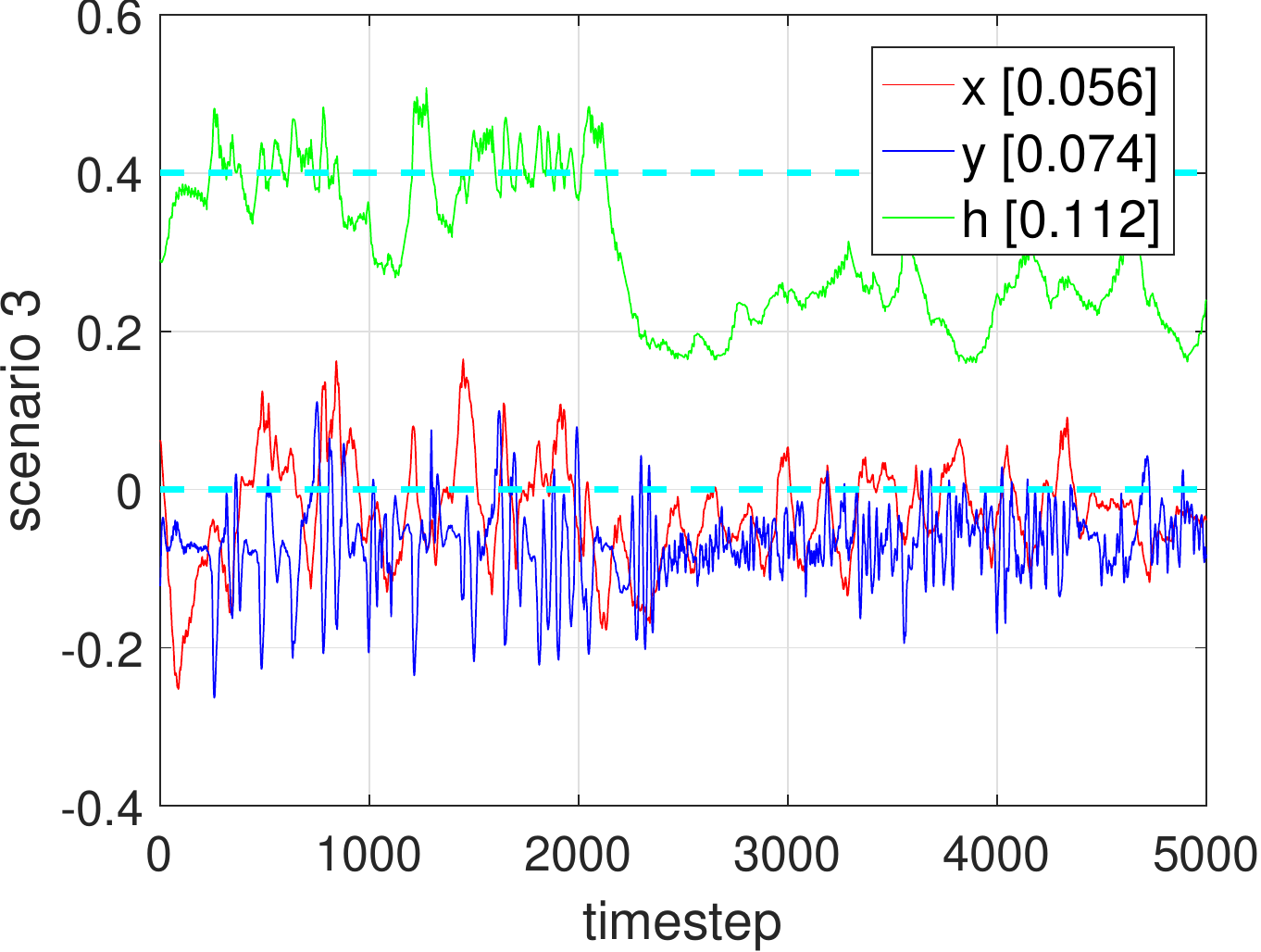}}
	\caption{Variations of the normalized true target state  $\s_{o,t} = (x, y, h)$ over time ($x,y\in[-0.5,0.5], h\in[0,1]$) in simulator evaluation. The dashed lines show the goal for each states: $(x^*,y^*,h^*) = (0, 0, 0.5)$.  The top-right legend shows the average deviation.}
\end{figure*}

We first compare two different policies trained by separate learning and end-to-end learning (as described in Figure~\ref{fig:2_comparison}), respectively.
By applying the models in the testing environment, we can record the true target state variation, as shown in Figure~\ref{fig:test_state}, to show the quality of different policies. We also compute the average deviation of each state variable to give quantitative analysis, as shown in the top-right legend. The result is consistent with our findings in the design evaluation. The end-to-end learning approach is clearly better.
It is worth noting that, although the policies are only trained with a maximum of 1000 time steps, the agent can generalize well beyond that.
Both policies can consistently follow the target for quite a long time, neither crashing the quadrotor nor losing the target from the camera view.
However, the separate learning approach only manages to learn a suboptimal policy with which the resulting target state is relatively far from the desired goal state, as indicated by the dashed line in Figure~\ref{fig:test_state}. On the contrary, the end-to-end learning approach achieves a very stable policy to successfully maintain the target state within a small range of the goal state for up to 5000 time steps (and even more). For all subsequent experiments in this section we will stick to the model trained by the end-to-end learning approach.

Since training is all done in scenario~1 (Figure~\ref{fig:target_task}), we further test the trained policy network in two unseen environments (scenarios~2 and~3) to evaluate the generalization ability. Scenario 2 has a background significantly different from the training setting and scenario 3 has very similar object distractors.
We directly apply the trained policy network to the new testing environments without any adaptation. As shown in Figure~\ref{fig:test_state_unseen}, surprisingly, our policy network exhibits good generalization ability to unseen scenes. The perception module demonstrates moderate tolerance to different backgrounds and distractors and the control module learns general goal-driven strategies. We find that in scenario 3 the scale drifts a little bit. This is the result of occlusion by some similar distractors.

\begin{figure}[htb]
	\centering
	\includegraphics[width=0.24\textwidth]{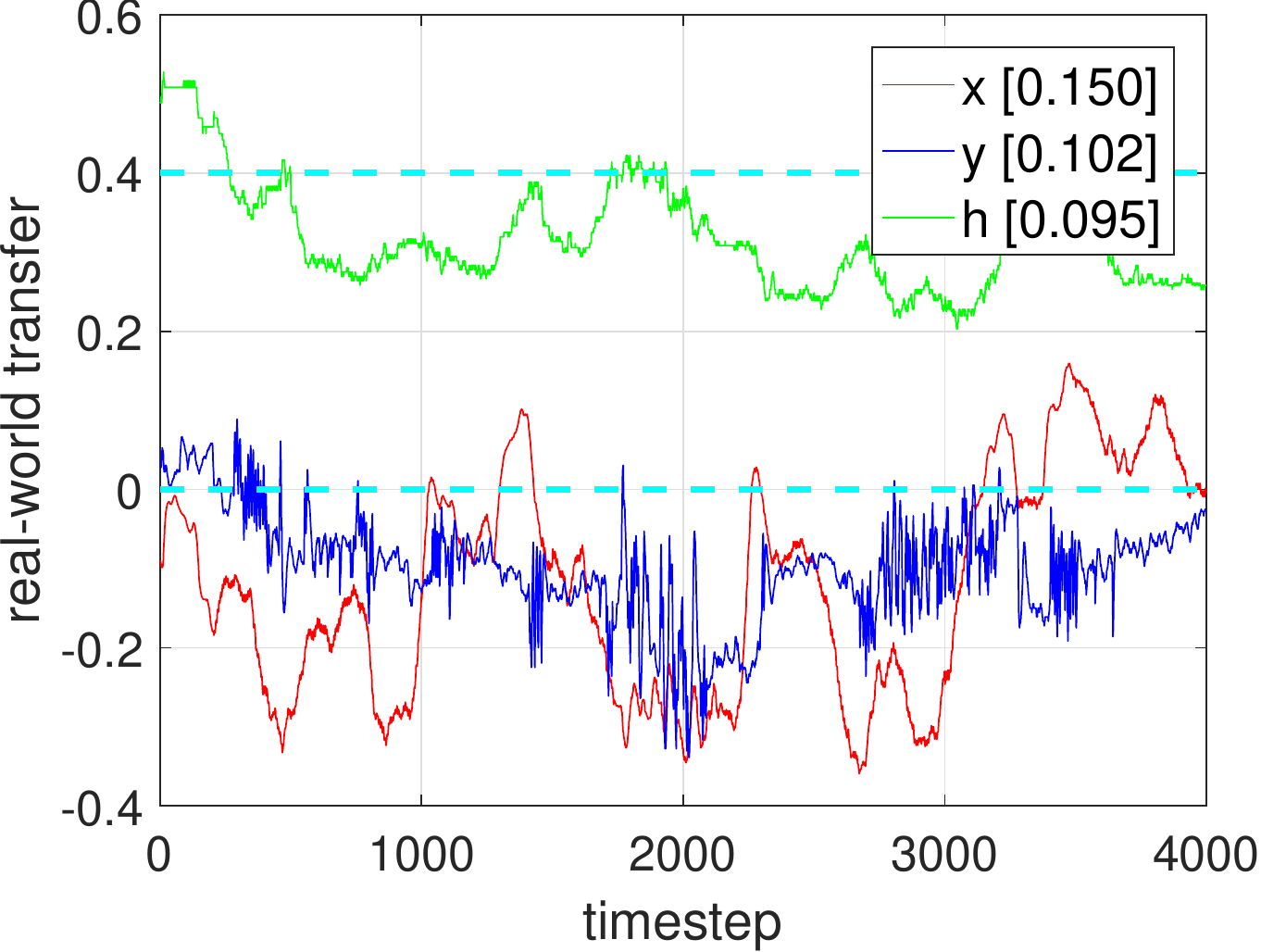}
	\caption{\label{fig:test_state_m100}Variations of the normalized true target state  $\s_{o,t} = (x, y, h)$ over time ($x,y\in[-0.5,0.5], h\in[0,1]$) in real world evaluation. The dashed lines show the goal for each states: $(x^*,y^*,h^*) = (0, 0, 0.5)$.  The top-right legend shows the average deviation.}
\end{figure}

\subsection{Policy Transfer to the Real World}
\label{sec:policy_transfer}
Our quadrotor testbed is based on the DJI Matrice 100 platform equipped with an Intel NUC and a camera. We use DJI built-in functions to map the high-level actions to actual flight commands, which are significantly more complex than a simple PID controller. This mapping computation is done by the NUC.  For speed consideration, the policy network computation is deployed on a ground laptop M1000 GPU which communicates with the onboard NUC by Robot Operating System (ROS).

Figure~\ref{fig:test_state_m100} shows the results of the real-world flight test, where the true target states are labeled off-line by an object tracker. With a slight decrease in control precision, the transferred agent can still follow the moving target for up to 4000 time steps (approximately 3 minutes). It is worth noting that this real-world test bears several challenges, including the large gap between the low-level controllers, the data delay, and the state noise in inertial measurement unit (IMU) and GPS. Without any adaptation, the learned policy can still exhibit reasonable following capability. This shows that the policy does have well-generalized performance. Note that the aim of this real-world test is not to achieve optimum performance, but to demonstrate the possibility of transferring a model trained in a simulator domain to a real-world domain. With more mature simulators, we believe these are promising directions to pursue in robotics learning.
\section{Conclusion and Future Work}
In this paper, we have explored the potential of applying a machine learning approach to the challenging autonomous UAV control problem.
The policy is represented by a convolutional neural network which combines perception and control and thus can be trained end-to-end.
Instead of directly predicting the low-level motor commands, the policy network is designed to produce high-level actions. This enables both stable learning and good generalization ability to different environments.
Our training approach consists of supervised learning from raw images and reinforcement learning from games of self-play. This training decomposition greatly alleviates the instability of existing model-free policy gradient methods. Results from both simulated and real-world experiments show that our method can successfully perform the target following task with good generalization ability.

Our approach also bears some limitations. The perception module does not generalize well to dramatically different settings, such as long-term occlusion. We may rely on incorporating temporal information into the model to alleviate this problem. 
Currently the policy is directly transferred to the real-world quadrotor. This can also be used as an initialization scheme and the model can be further trained end-to-end~\cite{zhu2017target}. How to set up a game playing loop in real-world environments is a bottleneck.
We will pursue these research directions in our future work.

\bibliographystyle{IEEEtran}
\bibliography{IEEEabrv,icra18}

\end{document}